\documentclass[12pt]{article}

\usepackage[margin=1in]{geometry}
\usepackage{setspace}
\usepackage{amsmath}
\usepackage{booktabs}
\usepackage{graphicx}
\usepackage{caption}
\usepackage{threeparttable}
\usepackage[round]{natbib}
\usepackage[hidelinks]{hyperref}

\title{Reading Is Not Using:\\
Retrieval, Judgment, and the Design of AI Financial Research Workflows\thanks{%
Very preliminary draft. Comments welcome. All experimental materials and code
are available upon request from the authors and we expect to release the
replication package around the end of September. We
thank the Columbia Business School Research Grid for providing computational
support for this work.}}
\author{%
Miao Liu\thanks{Carroll School of Management, Boston College;
\texttt{miao.liu@bc.edu}.}
\and
Zhizhe Liu\thanks{Columbia Business School, Columbia University;
\texttt{zhizhe.liu@columbia.edu}.}
}
\date{This draft: August 2026}

\begin{document}

\begin{titlepage}
\maketitle
\thispagestyle{empty}
\begin{abstract}
\noindent
Large language models (LLMs) are increasingly deployed as AI analysts to process financial disclosures and support AI-assisted investment decisions. Yet such systems are usually evaluated by what they can retrieve, not whether retrieved information affects their judgments. We identify a retrieval--integration gap in long-context financial analysis. Holding focal-firm information fixed and varying only unrelated context from 2,000 to 128,000 tokens, we find that a risk disclosure's influence on investment judgments falls to the experimental noise floor even as direct retrieval remains accurate. The pattern replicates across model families and judgment tasks and in experiments removing real disclosures from actual 10-K filings. More capable models postpone but do not eliminate the gap. Causal memory interventions show that compressed summaries and source-text lookup jointly transmit disclosures into judgments. Workflow architecture determines whether this transmission succeeds: chunk-and-summarize pipelines evict relevant information, whereas a targeted, structured restatement adjacent to the decision restores its influence. AI analyst performance is therefore jointly determined by model capability and workflow architecture. Retrieval-based evaluations can certify systems whose investment judgments ignore information they demonstrably retrieved.

\end{abstract}
\vspace{0.5em} \noindent\textbf{Keywords:} Generative AI; large language models; AI analysts; AI-assisted decision making; long-context analysis; AI workflow architecture; retrieval--integration gap.

\end{titlepage}

\doublespacing

\section{Introduction}\label{sec:intro}

Large language models (LLMs) are becoming \emph{AI analysts}: they read financial disclosures, answer firm-specific questions, and increasingly support investment decisions. Retail investors query generative AI tools about firms at scale, and aggregate trading activity is associated with the recommendations those tools provide \citep{ecker2026stock}. Capital-market intermediaries use AI to help produce investment research \citep{bradshaw2026generative}, while the technology is reshaping how information is produced, disseminated, and processed throughout capital markets \citep{cao2026generativeai}. Experimental evidence further shows that interacting with these tools changes investor judgments, and not always for the better \citep{croom2026interactivity}. By transferring both information acquisition and preliminary analysis to the model, users are doing more than adopting a passive tool: they are delegating a consequential task to an agentic information system \citep{baird2021next}. As LLMs move from document search into AI-assisted decision making, a basic performance question follows: does information that an AI analyst can retrieve actually influence the investment judgment it produces?

Finding information and using it are distinct stages of decision support. The disclosure processing cost framework separates the cost of locating information from the cost of integrating it into a judgment \citep{blankespoor2020disclosure}. Prior evidence similarly shows that reducing information-acquisition costs can change firms' disclosure choices without eliminating downstream processing frictions \citep{blankespoor2019xbrl}. LLMs sharply reduce the retrieval cost: when asked a targeted question, a modern model can often recite a covenant threshold or settlement amount from a 100,000-token filing. But retrieval accuracy does not establish decision usefulness. We call the divergence between what an AI analyst can retrieve and what affects its judgment the \emph{retrieval--integration gap}. We ask whether this gap widens as context grows and whether AI analyst performance depends not only on model capability but also on the architecture of the workflow connecting disclosure reading to investment judgment. This framing treats performance as an outcome of the complete AI system, rather than as an inherent property of the underlying model alone.

We identify the retrieval--integration gap through controlled experiments that measure retrieval and decision influence separately. For each of twelve U.S. registrants, we construct a firm-specific, quantitatively verifiable risk disclosure---such as a covenant threshold, settlement payment, or indemnification cap---that is coherent with the surrounding filing. We define the disclosure's \emph{marginal decision influence} as the difference between the model's investment judgment when the disclosure is present and its average judgment when the same location contains one of five equal-length neutral replacements. We then expand only the economically unrelated, genre-matched surrounding text from 2,000 to 128,000 tokens, leaving the decision-relevant information about the focal firm unchanged. This design isolates the effect of context load from legitimate updating on additional firm information. In separate calls scored against frozen answer keys, we test whether the model can still retrieve the disclosure's facts. The resulting measures distinguish what an AI analyst can state from what it actually uses. Section~\ref{sec:design} provides the complete design, and Table~\ref{tab:design} summarizes it.

\begin{figure}[p]
\centering
\includegraphics[width=0.7\textwidth]{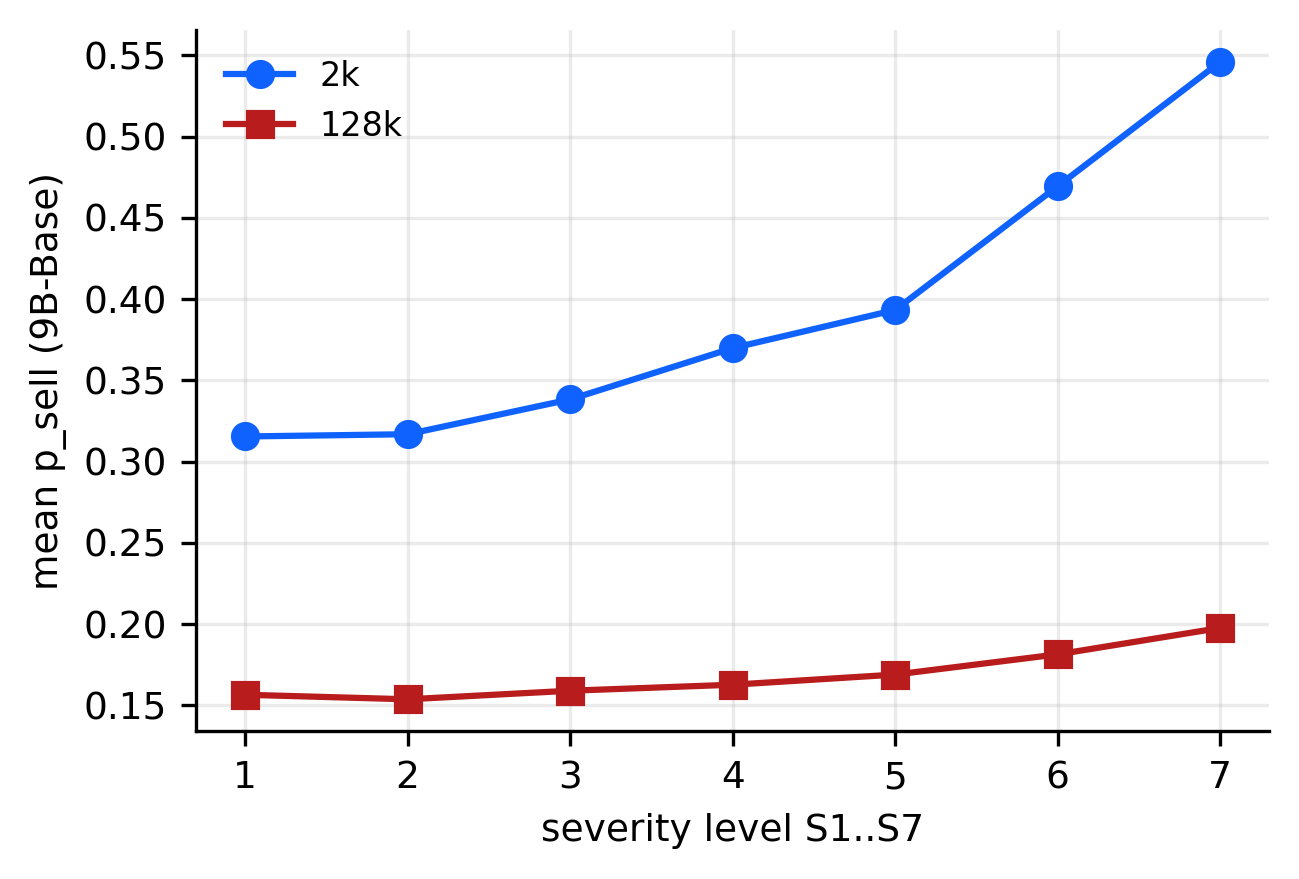}
\caption{\textbf{Ordinal sensitivity survives; cardinal transmission
collapses.} Mean sell probability by disclosure severity (S1--S7) at 2k and
128k tokens (primary model, 12 firms). The model's response range across
the full severity ladder compresses from 0.23 to 0.04---a factor of
5.6---while its ability to \emph{order} severities only partly degrades
(pairwise ordering accuracy 0.93 $\rightarrow$ 0.79). At full length the
model can still rank risks; it no longer scales its judgment to them.}
\label{fig:severity}
\end{figure}

\begin{figure}[p]
\centering
\includegraphics[width=\textwidth]{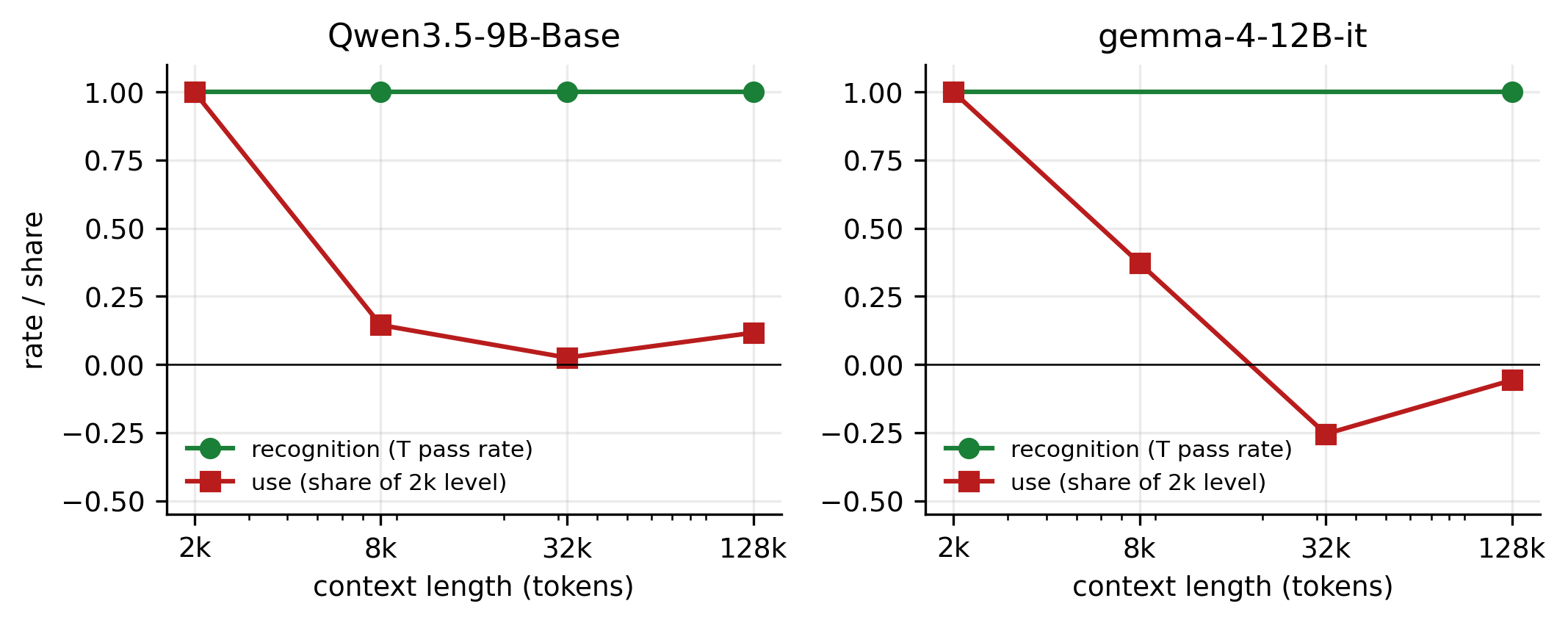}
\caption{\textbf{Reading survives; using dies.} Directed retrieval of the
target facts (green) remains at ceiling at every context length, while the
same disclosure's influence on the investment judgment (red, scaled by its
own 2k level) collapses --- and in gemma-4 mildly inverts.}
\label{fig:reading_using}
\end{figure}

Integration does not keep pace with retrieval. For our primary open-weight model, the disclosure raises the model's sell probability by 3.2 percentage points in a 2,000-token filing---an economically meaningful response to one paragraph. Between 8,000 and 32,000 tokens, however, its influence becomes statistically indistinguishable from the effects of inserting neutral text and remains at that empirical noise floor through 128,000 tokens (Figure~\ref{fig:h1}; Table~\ref{tab:main_use}). A higher-capability production model postpones but does not avoid the decline: its response falls by roughly half and is no longer distinguishable from neutral-insertion noise at full length. Direct retrieval, by contrast, remains stable. At 128,000 tokens, the primary model retrieves the disclosure for all twelve firms and produces no false retrievals on neutral filings; retrieval by the higher-capability production model is similarly undiminished (Table~\ref{tab:recognition}; Figure~\ref{fig:reading_using}). Nor does context length simply make judgments uniformly less responsive. Across a seven-step severity ladder, the model's response range compresses by a factor of 5.6 while its ability to order the risks degrades much less (Figure~\ref{fig:severity}). At full length, the model can still rank risks but no longer scales its judgment to their severity. Direct factual accuracy and downstream decision influence therefore separate sharply: reading is not using.

\begin{figure}[p]
\centering
\includegraphics[width=0.7\textwidth]{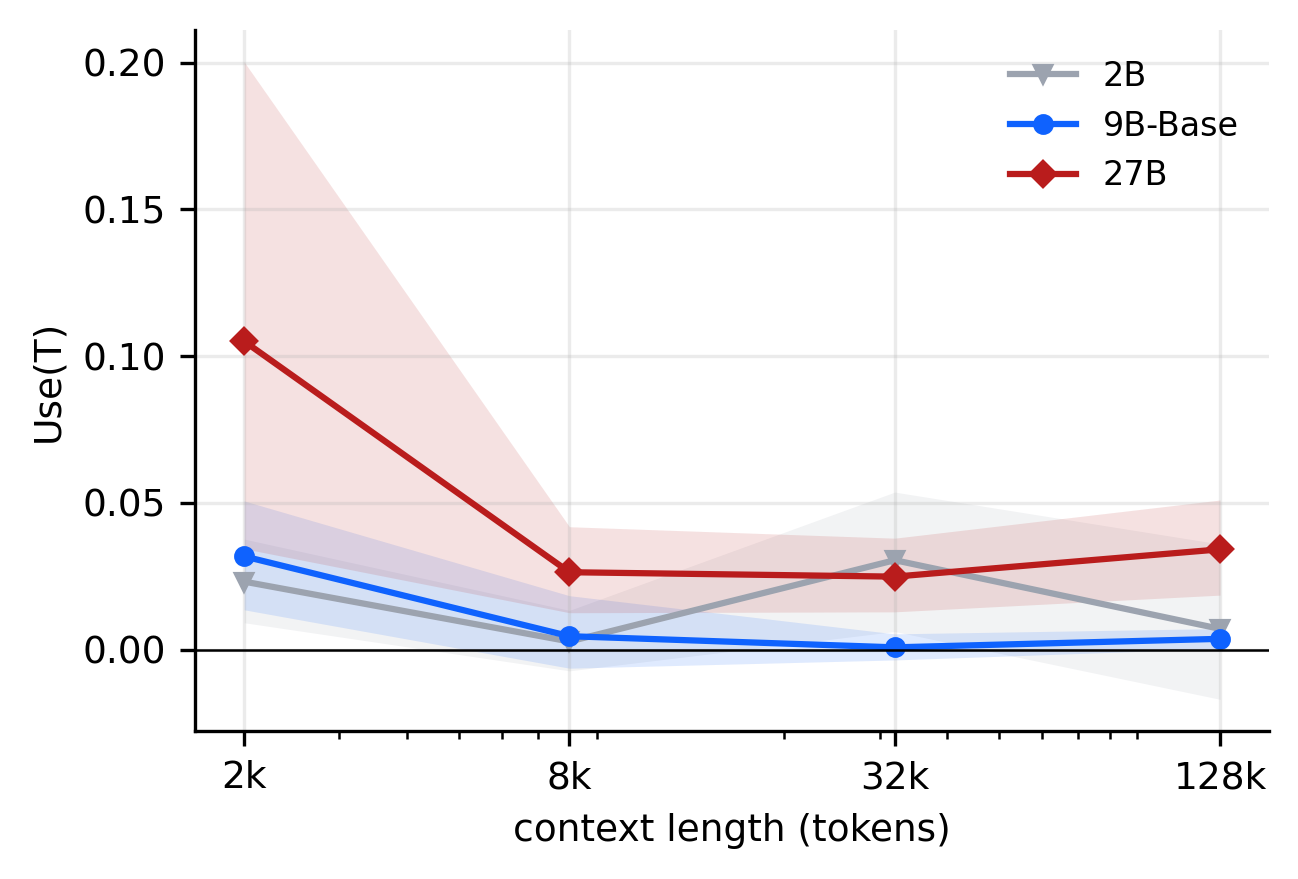}
\caption{\textbf{Capability moves the integration horizon outward.} Within
the Qwen3.5 family, the largest model is the only one whose use of the
disclosure remains statistically distinguishable from zero at every
length, retaining at 128k tokens roughly what the mid-sized model achieves
at 2k; the ordering is not a monotone size ladder cell by cell.}
\label{fig:scale}
\end{figure}

The result is not specific to one model family or to constructed disclosures. The retrieval--integration gap replicates across three independently trained model families, although the context length at which it becomes binding differs across them. It also appears in an exploratory experiment using real disclosures. We identify twenty complete 10-K filings containing self-contained, quantified disclosures and compare the model's judgment before and after removing the original passage. The real disclosure affects judgment in a short excerpt but has essentially no influence in the complete filing, even though the model retrieves it correctly in all twenty cases (Table~\ref{tab:real_disclosure}). Capability shifts the failure frontier outward. The largest open-weight model retains a 3.4-percentage-point effect at 128,000 tokens---approximately what the mid-sized model achieves at 2,000 tokens---whereas a lower-capability commercial system eventually loses retrieval as well as decision influence (Figure~\ref{fig:scale}; Table~\ref{tab:scale}). Capability therefore buys distance from the failure, not evidence of immunity from it: the effective decision context of an AI analyst need not equal its advertised context window.

We next investigate why retrieval and decision influence diverge. Because the processor is a model rather than a person, we can inspect its intermediate representations and intervene directly on the channels through which previously read information reaches the final judgment. Representation analyses first show that the disclosure's risk content remains stably encoded where it was read at every context length. The model therefore forms a usable representation of the disclosure even in long documents. At the decision position, however, standard probes do not recover disclosure-specific content, and the legibility of the pending decision deteriorates as context grows (Section~\ref{sec:mechanism}; Figure~\ref{fig:mechanism}).

The architecture provides two potential transmission channels from reading to judgment: a compressed recurrent state that summarizes preceding content and attention-based lookup over the source text. Causal interventions isolate both channels. Transplanting the compressed state between matched documents removes roughly two-thirds of the disclosure's influence when the disclosure is erased from that state, even though its source text remains available for lookup. Conversely, implanting the disclosure into the compressed state recreates nearly half of its influence in a document that does not contain it (Table~\ref{tab:transplant}). Matched attention-blackout experiments identify a disclosure-specific lookup effect of comparable magnitude, and the two estimated channel effects are statistically indistinguishable. These interventions locate the failure at integration rather than comprehension: the disclosure remains encoded and retrievable, but the channels carrying it into the judgment weaken as competing context grows.

The mechanism generates a direct system-design prediction: reorganizing information flow should be more effective than merely increasing computation. The workflow experiments support that prediction (Section~\ref{sec:remedies}; Table~\ref{tab:remedies}). Enabling extended reasoning does not restore the disclosure's influence and reduces it at short context lengths. A generic chunk-and-summarize workflow performs even worse: it eliminates the disclosure's influence at every length, including 2,000 tokens, because its bounded notes omit the target information before the decision stage sees it. Simply repeating the disclosure verbatim near the decision also accomplishes little. What works is a targeted, structured restatement of the disclosure's decision-relevant facts placed immediately before the investment judgment while the source filing remains available. This workflow raises the disclosure's influence at 128,000 tokens to 8.5 percentage points, with all twelve firms responding in the predicted direction. An experimenter-written restatement works as well as the model's own extraction, and the remedy also succeeds in complete filings containing real disclosures (Tables~\ref{tab:decomposition} and~\ref{tab:eco_remedies}). The same model, information, and decision task therefore produce materially different judgments solely because the workflow routes information differently. AI analyst performance is jointly determined by model capability and workflow architecture; retrieval-based evaluations can certify systems whose investment judgments ignore information they demonstrably retrieved.

This paper makes four contributions. First, it extends IS research on delegation from the allocation of work between humans and AI to the execution of work within an agentic artifact. Prior research develops theory and evidence on how tasks and decision rights should be allocated between humans and AI and when alternative arrangements improve performance \citep{baird2021next,fugener2022cognitive}. We study the next stage: what happens after an end-to-end task has been delegated and the artifact must carry information across its own processing stages. We identify a distinct post-delegation failure: an AI agent can demonstrably retrieve decision-relevant information yet fail to incorporate it into the judgment it was delegated to form. The retrieval--integration gap thus shows that successful performance on a verifiable intermediate operation need not imply completion of the delegated objective---a failure that becomes increasingly consequential as agentic systems perform multistage tasks with limited human visibility.

Second, we contribute a task-aligned approach to the design and evaluation of generative AI systems. Technically grounded IS design research calls for rigorous artifact evaluation, consequential domain adaptation, and actionable design knowledge \citep{abbasi2024pathways}. Related research identifies a ``construct gap'' when the observable proxy around which an ML system is designed omits important dimensions of the decision-relevant construct \citep{dearteaga2025leveraging}. We identify an analogous gap in AI evaluation: retrieval accuracy is a convenient and observable performance proxy, but it does not establish that retrieved information affects the decision the system is meant to support. To measure that ultimate objective, we introduce \emph{marginal decision influence}, a continuous counterfactual measure of how much a particular input changes an AI system's judgment while all other decision-relevant information is held fixed. The measure reveals that retrieval-based evaluations can certify systems whose downstream judgments are effectively invariant to the facts they successfully retrieve.

Third, we move from detecting the retrieval--integration gap to explaining it and deriving design knowledge. Internal interventions identify two channels through which a disclosure can reach judgment---a compressed running summary and attention-based lookup---and show how context load disrupts the transmission of decision-relevant information. The workflow experiments yield a decision-proximal representation principle: information critical to a judgment should be extracted into a targeted, structured representation and placed at the point of judgment rather than routed through generic, capacity-constrained summaries. Whereas \citet{chen2024large} show that the consequences of human--LLM collaboration depend on collaboration modality and user expertise, we extend this configuration logic to the internal workflow of a delegated AI analyst. Generic chunk-and-summarize workflows and additional reasoning fail to restore the disclosure's influence, whereas a targeted, structured restatement succeeds. Consistent with calls to manage AI as an organizationally embedded and partly inscrutable artifact \citep{berente2021managing}, our results show that model capability and workflow architecture play distinct but complementary roles: capability moves the failure frontier, whereas workflow architecture determines whether decision-relevant information reaches the judgment at all.

Fourth, we contribute to research on financial disclosure processing and machine readership by distinguishing \emph{machine readability} from \emph{machine decision usefulness}. Generative AI does not eliminate disclosure processing costs; it relocates them from finding information to integrating it into judgment \citep{blankespoor2020disclosure}. Retrieval becomes inexpensive, but integration remains sensitive to document length, model capability, and workflow architecture. For investors and intermediaries, this distinction means that an AI analyst should be validated by whether economically meaningful changes in its information produce appropriate changes in its judgment, not merely by whether it answers factual questions correctly. For disclosure regulators, machine-readable presentation and successful retrieval do not establish that a disclosure is decision useful to a machine reader. More broadly, the financial-disclosure setting demonstrates how the information artifact, the model, and the surrounding workflow jointly determine whether available information is transformed into consequential action.

Our mechanism claims are deliberately bounded. We study AI processors, not human cognition, and therefore do not examine whether users trust, accept, or override the resulting recommendations. The connection to the limited-attention literature, in which salience and processing load affect market responses even when information is public \citep{hirshleifer2003limited,hirshleifer2009driven}, is an economic analogy rather than a claim that humans and LLMs share a cognitive mechanism. Our analysis instead isolates the reliability of the artifact after a user has delegated the analytical task to it.

Relative to computer-science research on long-context degradation, which shows that additional context can reduce retrieval and reasoning accuracy even when relevant information remains accessible \citep{liu2024lost,levy2024same,du2025context}, our contribution lies in measuring an economically consequential downstream judgment, holding the focal information set fixed, causally identifying the channels that transmit a retrieved fact into that judgment, and testing workflow remedies. Relative to emerging accounting and finance research on what LLMs can produce from financial statements and market information \citep{lopezlira2026chatgpt,dekok2025chatgpt,cao2026generativeai}, we ask whether a particular disclosure changes the model's judgment. That is the stage at which an AI analyst's output becomes economically consequential and a disclosure becomes decision useful.

The remainder of the paper proceeds as follows. Section~\ref{sec:background} develops the framework and hypotheses. Section~\ref{sec:design} presents the experimental design. Section~\ref{sec:phenomenon} establishes the retrieval--integration gap and its robustness. Section~\ref{sec:mechanism} identifies the transmission mechanism, and Section~\ref{sec:remedies} evaluates alternative workflow architectures. Section~\ref{sec:conclusion} concludes.

\section{Background and Hypotheses}
\label{sec:background}

\subsection{Delegation to an Agentic IS Artifact}

When an investor asks a language model to read a filing and recommend an
action, the model is not being used as a passive tool. The user transfers
both the acquisition of information and the analytical work that follows,
and receives back a judgment rather than a document. Information systems
research characterizes this arrangement as delegation to an agentic IS
artifact: the artifact holds decision rights over how a task is carried out,
and the user's role shifts from performing the work to specifying it and
accepting or rejecting its output \citep{baird2021next}. A parallel stream
studies how tasks should be allocated between humans and machines and shows
that delegation improves performance only under particular conditions, since
people delegate poorly when they cannot observe where the machine is strong
\citep{fugener2022cognitive}. Both streams take the boundary of delegation as
the object of theory: which party should do what, and with what visibility
into the other's competence.

This paper takes up what happens on the far side of that boundary. Once an
end-to-end analytical task has been delegated, the artifact must execute it
internally, carrying information across its own processing stages---reading
the document, forming an internal representation of what it read, and
producing a judgment conditioned on that representation. Each stage is a
transmission step, and none of them is observable to the user, who sees only
the final recommendation. Delegation therefore creates a distinctive failure
mode that is invisible at the delegation boundary: the artifact may complete
a verifiable intermediate operation and still fail to complete the objective
it was delegated to achieve. That divergence between what an AI analyst can
retrieve from a document and what actually affects the judgment it
produces---the \emph{retrieval--integration gap}---is what this paper
measures.

The gap is consequential precisely because agentic artifacts are trusted on
the strength of intermediate evidence. A user who watches a model quote a
covenant threshold verbatim from a hundred-page filing has direct evidence
that the information was found, and reasonably infers that the subsequent
recommendation reflects it. That inference is what we test. Managing AI as an
organizationally embedded and partly inscrutable artifact requires knowing
which observable behaviors license which conclusions about unobservable ones
\citep{berente2021managing}, and the retrieval--integration gap is a case in
which a highly legible behavior licenses much less than it appears to.

\subsection{Why Disclosure Processing Makes the Distinction Consequential}

Financial disclosure is a setting in which the difference between finding
information and using it has long been theorized and has direct economic
stakes. A large literature organizes the frictions between mandated
disclosure and investor behavior around processing costs: before a disclosure
can move a decision, an investor must become aware that it exists, acquire
it, and integrate it into a judgment \citep{blankespoor2020disclosure}. The
framework's central move---treating acquisition and integration as separate
costs that need not fall together---is exactly the distinction that the
retrieval--integration gap makes operational for machine readers. Each stage
is costly, and the costs are not small. Annual reports have grown longer and
less readable over time \citep{li2008annual, loughran2014measuring}, with
much of the growth concentrated in boilerplate-heavy sections that bury
firm-specific content \citep{dyer2017evolution}. Market participants respond
to these costs the way the framework predicts: when a mandate lowered users'
processing costs, firms supplied more detailed disclosure in response
\citep{blankespoor2019xbrl}, and when machine readership of filings became
widespread, firms adjusted how they write to accommodate the machines that
now read them first \citep{cao2023talk}.

Generative artificial intelligence changes the terms of this bargain more
abruptly than any prior processing technology. A large language model can
take an entire 10-K in a single pass and, on direct questioning, locate a
footnote, quote an amount, and characterize a risk in seconds. Early evidence
suggests such capabilities translate into economically meaningful analysis:
language models extract decision-relevant information from financial
disclosures \citep{huang2023finbert, dekok2025chatgpt}, extract
return-relevant signals from news \citep{lopezlira2026chatgpt}, predict
future earnings changes when fine-tuned on financial statements
\citep{dou2026supervised}, and machine analysis of corporate disclosures can
rival human analysts' forecasts \citep{cao2024man}. Adoption on the user side
is no longer hypothetical. Retail investors interact with AI assistants about
specific stocks at scale \citep{ecker2026stock}, capital-market
intermediaries visibly use generative tools in their published research
\citep{bradshaw2026generative}, sell-side analysts adopt them in forecasting
\citep{xue2025generative}, and the profession is beginning to take stock of
how AI reshapes information production and processing end to end
\citep{cao2026generativeai}.

Read through the delegation lens, the processing-cost framework yields a
sharper question than whether these tools are useful. AI changes the
acquisition stage most visibly: once a document is in context, a capable
model locates and recites a named item almost without error. (Our design
measures exactly this conditional retrieval---of an item the question
names---not awareness, and not the time or money costs of processing.) But
the framework treats integration as a separate cost, and there is no a priori
reason it falls in proportion. Indeed, experimental evidence on human users
already hints at a wedge between access and judgment: interacting with a
generative AI assistant raises investors' confidence in their information
advantage more than it raises their actual judgment quality
\citep{croom2026interactivity}. This paper asks the analogous question about
the artifact itself. When a model demonstrably can retrieve a
disclosure---when directed retrieval of the named item is essentially
perfect---does the disclosure actually enter the model's judgment? If
retrieval becomes inexpensive while integration does not, the binding
constraint on disclosure effectiveness moves from \emph{finding the
information} to \emph{finding it and not using it}. That distinction
separates machine readability from machine decision usefulness, and it is a
failure mode that standard evaluations of AI reading comprehension are not
designed to detect.

\subsection{Evaluating a Delegated Decision Task}

The evaluation problem is not unique to disclosure. Technically grounded IS
design research argues that generative systems should be assessed by rigorous
artifact evaluation against consequential domain outcomes rather than by
convenient benchmark performance \citep{abbasi2024pathways}. Research on
machine-learning decision support identifies the underlying difficulty as a
``construct gap'': systems are optimized and validated against an observable
proxy that omits dimensions of the construct the decision actually turns on
\citep{dearteaga2025leveraging}. Retrieval accuracy is such a proxy. It is
cheap to score, it is unambiguous, and it certifies that the artifact
completed one processing stage---but it is silent on whether the retrieved
content reached the judgment the artifact was delegated to form. Closing that
gap requires an outcome measure defined on the delegated task itself. We use
the \emph{marginal decision influence} of a disclosure: the change in the
artifact's judgment attributable to the presence of that disclosure, holding
all other decision-relevant information fixed. It is continuous rather than
binary, defined against an explicit counterfactual, and it moves only when
information changes the decision.

Computer science has documented that long inputs degrade language-model
performance, and this literature supplies the proximate reason to expect
transmission across processing stages to fail. Models answer questions less
accurately when relevant material sits in the middle of a long context
\citep{liu2024lost}; reasoning accuracy falls as inputs lengthen even when
the added text is irrelevant \citep{levy2024same}; and effective context
windows measured on controlled benchmarks are far shorter than advertised
ones \citep{hsieh2024ruler, yen2025helmet}. The result closest to ours is
that of \citet{du2025context}, who show that input length alone reduces
downstream task performance even when retrieval of the relevant passage is
perfect. Mechanistic accounts point to attention: a small set of attention
heads carries long-context recall \citep{wu2025retrieval}, and positional
biases in attention explain much of the middle-of-context loss
\citep{hsieh2024found}.

Three things are missing from this literature for evaluating a delegated
decision task. First, the dependent variable. Benchmark studies score answers
right or wrong, which is precisely the proxy whose sufficiency is in
question; assessing a delegated judgment requires measuring whether retrieved
information \emph{changes} that judgment---a continuous quantity, measured
against an explicit counterfactual in which the information is absent.
Second, identification. Comparing a short excerpt with a full filing
confounds context length with the information the longer document adds; a
clean answer requires holding the focal firm's information set fixed while
varying only the surrounding load. Third, consequences and design knowledge.
If long-context degradation afflicts delegated financial analysis, the
practical question is which processing arrangements restore the lost
influence---a question about workflow architecture, and one whose answer is
actionable for anyone assembling an AI analyst from a model, a document
store, and a sequence of intermediate steps.

A separate lesson from the interpretability literature disciplines our
mechanism analysis. That information can be decoded from a model's internal
states does not imply the model uses it: a probe can recover content that
plays no causal role in the output, and even interventions on
plausible-looking internal directions can produce behavioral effects for
uninteresting reasons \citep{makelov2024subspace}. We therefore treat
``the model read it'' and
``the model used it'' as distinct, separately measured claims throughout,
and we hold internal-representation evidence to explicit control and
resolution standards described in Section~\ref{sec:mechanism}.

\subsection{Hypotheses}

Our first hypothesis states the retrieval--integration gap as a comparative
prediction and translates it into the fixed-information design of
Section~\ref{sec:design}.

\medskip
\noindent\textbf{H1.} \emph{Holding the focal firm's information set fixed,
the marginal influence of a risk disclosure on the model's investment
judgment declines as unrelated context grows.}
\medskip

The reasoning is about transmission under load. A delegated analytical task
requires the artifact to carry a specific fact from the point where it was
read to the point where the judgment is formed, and every additional token of
unrelated material competes for the finite representational capacity that
carries it. The economic analogue is limited attention: investors underreact
to relevant news when competing stimuli crowd the processing stage
\citep{hirshleifer2009driven}, and disclosure placement and salience matter
precisely because attention, not access, binds \citep{hirshleifer2003limited}.
H1 asks whether an artifact with no fatigue and every word of the filing
simultaneously in view exhibits the same signature. We draw the analogy at
the level of the prediction, not the mechanism.

\medskip
\noindent\textbf{H2.} \emph{The decline in influence is not retrieval
failure: at context lengths where the disclosure's influence on judgment has
fallen, the model still retrieves the disclosure accurately under direct
questioning.}
\medskip

H2 is what makes the failure a post-delegation failure rather than a reading
deficit. If retrieval and influence fell together, longer contexts would
simply make models worse readers, the artifact would be failing at an
observable stage, and the remedy would be better retrieval. If retrieval
survives while influence dies, the artifact completes the stage its evaluator
can see and fails at the stage its evaluator cannot---the integration stage
that the processing-cost framework identifies as distinct. H2 is thus the
test of whether the retrieval proxy suffers the construct gap described
above: a system passing it may nonetheless produce judgments invariant to the
facts it demonstrably retrieved.

\medskip
\noindent\textbf{H3.} \emph{If within-context retention is the operative
mechanism, text arriving after the disclosure should reduce its influence
more than the same text arriving before it.}
\medskip

H3 localizes where in the artifact's processing the transmission degrades.
Placing filler after the target increases the distance the disclosure must
travel to the point of decision; placing it before does not. Architectural
considerations make this comparison informative: models with recurrent or
compressed sequence processing exhibit measurable limits on carrying early
content forward \citep{jelassi2024repeat, arora2024zoology}. We designate H3
as a pre-specified secondary comparison rather than a primary test, because a
null behavioral difference would not rule out retention failures that
internal measurements can still detect.

\medskip
\noindent\textbf{H4.} \emph{Reorganizing the decision process restores the
disclosure's influence at long context; adding computation does not.}
\medskip

H4 is a hypothesis about workflow architecture rather than about the model.
If the disclosure is present, retrievable, and nonetheless unused, the
deficiency lies in how the system routes information between its stages, not
in how much capability or computation it applies at any one of them.
Reorganizing that routing---for example, requiring the artifact to state what
the filing says about the risk immediately before deciding---should recover
influence, whereas allowing the model to deliberate longer should not, and a
generic summarization stage that compresses the document before the decision
stage sees it may make matters worse. The prediction extends a configuration
logic already established for human--AI collaboration, where outcomes depend
on how the collaboration is arranged rather than on the model alone
\citep{chen2024large}, to the internal arrangement of a delegated AI analyst.
H4 turns the mechanism into treatments: each remedy in
Section~\ref{sec:remedies} corresponds to a distinct hypothesis about where
the integration failure sits, their comparative performance is itself
evidence about mechanism, and the comparison yields design knowledge about
where decision-relevant information should be represented in an agentic
workflow.

\section{Experimental Design}
\label{sec:design}

Table~\ref{tab:design} summarizes the design. Its purpose is to measure one
quantity cleanly: the marginal influence of a single, known risk disclosure
on an AI analyst's investment judgment, as a function of how much other text
surrounds it. The design measures retrieval and decision influence
separately, in distinct calls, so that what the AI analyst can state about a
disclosure and what that disclosure does to its judgment are two independent
observations rather than one. That separation is what renders the
retrieval--integration gap observable.

\subsection{The Measurand}

\emph{Marginal decision influence} is a continuous, counterfactual measure of
how much one input changes an AI system's judgment while all other
decision-relevant information is held fixed. It is defined on the delegated
decision task itself rather than on a retrieval proxy: the quantity of
interest is movement in the judgment the AI analyst was asked to produce, not
accuracy on an intermediate question about the document. We construct it as
follows.

For each firm and context length $\ell$, we construct a family of otherwise
identical filings that differ only in one paragraph. The \emph{target}
version ($T$) contains a firm-specific risk disclosure. Five \emph{neutral}
versions ($N_1,\dots,N_5$) replace that paragraph with equal-length neutral
text. The marginal decision influence of the disclosure is
\begin{equation*}
\mathrm{Use}(\ell) \;=\; \mathrm{Decision}(T,\ell) \;-\;
\frac{1}{5}\sum_{m=1}^{5}\mathrm{Decision}(N_m,\ell),
\end{equation*}
the difference between the model's judgment with the disclosure present and
its average judgment across the neutral replacements. Averaging over five
replacements matters because any single replacement text induces its own
deterministic shift in a model's output; the average absorbs those
idiosyncrasies so that $\mathrm{Use}(\ell)$ isolates the disclosure itself
rather than the accident of what replaced it. Because $\mathrm{Use}(\ell)$ is
a difference in the artifact's own decision output, it is expressed in the
units of the delegated task and remains well defined regardless of whether
the model can articulate the disclosure on request. That is the property a
retrieval score lacks, and it is why the two measures can diverge.

The decision readout differs by model class. For open-weight models we read
the model's propensity to recommend selling versus buying directly from its
output distribution at a fixed decision prompt---a deterministic, continuous
readout that requires no sampling. Commercial API models do not expose
output distributions, so for them we elicit a unit-bearing numeric judgment:
the probability the firm defaults within twelve months, in percentage
points. The two readouts are never mixed; each model's decline is measured
in its own units against its own baseline.

The sell-versus-buy propensity is not the only judgment we elicit. A
second task asks, of the identical documents, whether a credit-rating
downgrade within twelve months is likely, read out through the same
restricted one-token machinery so that the answer is again a
deterministic probability. Any conclusion that depended on how the
investment question happens to be phrased should fail to survive this
change of question; Section~\ref{sec:phenomenon} shows that the central
result survives it.

Separately---in a different call, by design---we measure \emph{retrieval}:
we ask the model directly what the filing says about the target matter and
score its answer against frozen answer keys covering the disclosed amounts
and status. Keeping the retrieval and decision readouts in separate calls is
deliberate. It means H2 compares two independent measurements, and no
instruction to ``first find, then use'' contaminates the decision task. The
resulting pair---one reading on the retrieval proxy, one on the delegated
decision---is what makes the retrieval--integration gap a measured quantity
rather than an inference from a single score.
Because both readouts are deterministic, differences across conditions are
not sampling noise; run-to-run variation is controlled by pinning hardware
and software configuration (Appendix~\ref{app:measurement}).

\subsection{Materials}

The experimental environment is built from the annual reports of twelve
U.S.\ registrants. For each firm we author a target risk paragraph---a
covenant threshold with limited headroom, a litigation settlement, an
indemnification cap, or a similar locally describable exposure---containing
verifiable amounts and written to be insertable at a specific location in
the notes or MD\&A. Authoring the paragraph, rather than sampling existing
disclosures, is what makes the counterfactual exact: we know precisely what
information the neutral versions remove.

The threat to validity from authored text is not whether it ``sounds
real''; it is whether the inserted fact is consistent with the rest of the
filing. Every candidate therefore passes a whole-filing coherence audit
against the financial statements, related notes, MD\&A, risk factors, legal
proceedings, and audit report, and against the filing's timeline. We impose
a single-edit rule: if making the fact true would require changes at more
than one location in the filing---restating a covenant table, altering a
subtotal---the candidate is discarded rather than patched. Multi-point
edits would turn one treatment into several. The audit protocol,
per-firm materials, and a seven-level severity ladder used to calibrate the
decision readout's sensitivity (from benign to near-default variants of
each disclosure) are described in Appendix~\ref{app:materials}.

\subsection{The Length Manipulation}

The identification problem with the natural comparison---a short excerpt
versus the full 10-K---is that the full filing adds information about the
same firm: mitigating detail, related exposures, context that could
legitimately dilute or amplify the target's weight. A rational reader
\emph{should} respond differently. Our main manipulation therefore holds the
focal firm's information set fixed. Each firm's material is condensed into a
fixed mini-filing containing the target (or neutral) paragraph, its local
section context, and the same decision prompt. Context is then extended to
2{,}000, 8{,}000, 32{,}000, and 128{,}000 tokens by adding text that is
economically unrelated to the firm and the target risk but matched in
genre---accounting and legal prose of the kind that populates real filings.
Nothing about the focal firm changes across lengths; only the surrounding
load does.

``Unrelated'' is a verified property, not an assumption. Before entering the
experiment, filler pools are certified by measuring their effect on
pure-filler baselines; filler text does shift the level of model judgments,
and the differenced design absorbs exactly such level shifts. Within each
length we also manipulate \emph{arrangement}: the filler appears either
entirely before the focal mini-filing or entirely after it, holding total
length and content fixed. The before/after comparison is the pre-specified
secondary test of H3.

A finer version of the same manipulation separates how much text the model
has read from how far the disclosure sits from the decision. Holding total
context fixed at 32{,}000 or 128{,}000 tokens, the filler is split around
the focal filing so that the disclosure's distance to the decision question
takes values from 1{,}000 to 64{,}000 tokens, while the union of filler
text remains byte-identical across the cells of each row. If proximity to
the decision is what preserves influence, moving the disclosure adjacent
to the question should restore it; if total load is what matters, distance
should be irrelevant.

Two extensions connect the controlled design to real reporting environments.
A same-genre \emph{semantic interference} arm replaces unrelated filler with
risk disclosures of other firms. And an \emph{ecological} arm inserts the
target paragraph into complete, investor-visible 10-K filings---with any
conflicting original passages removed through recorded, machine-verified
edits---accepting that this condition changes the firm's information set in
exchange for realism (Appendix~\ref{app:robustness}).

A third extension removes the authored paragraph from the design
altogether. For twenty additional registrants---every filing dated after
the models' training cutoffs---we identify a genuinely self-contained,
quantified disclosure already present in the filing: covenant headroom, a
litigation settlement, an indemnification obligation, a customer
concentration. Use is then measured by redaction rather than insertion:
the verbatim filing is compared with a version in which the disclosure is
surgically replaced by a length-matched neutral paragraph, and every other
passage that repeats the disclosure's distinctive amounts is excised from
\emph{both} versions, so the contrast isolates the disclosure itself. A
placebo edit---the same replacement applied to an unrelated, token-matched
passage---bounds the effect of editing per se; a window of roughly
2{,}000 tokens around the disclosure provides the short-context endpoint;
and directed retrieval on the redacted version doubles as a probe for
leakage and memorization, since a model that answers the retrieval
question without the disclosure must be drawing on something other than
the document. Materials and audits appear in Appendix~\ref{app:materials}.

\subsection{Inference}

The unit of observation is the firm-filing. Our design and primary
contrasts were fixed in advance of the frozen runs reported here; analyses
that go beyond them are labeled exploratory
where they appear. All formal statements rest
on two complementary procedures. First, a firm-level bootstrap resamples the
twelve firms with replacement to attach confidence intervals to mean
$\mathrm{Use}(\ell)$. Second, and more stringently, we construct an
\emph{empirical null}: for every cell we run ten additional
\emph{pseudo-target} insertions ($P_1,\dots,P_{10}$), which apply the
identical insertion machinery to paragraphs with no decision-relevant
content. These pseudo-effects are not zero---insertion itself perturbs a
deterministic reader---and they define the noise floor of the instrument.
Target effects are judged against this distribution by randomization
inference, drawing one pseudo-insertion per firm and recomputing the pooled
statistic.

The empirical null enforces a resolution principle that matters most at the
long end: a null result at 128{,}000 tokens is interpretable as a loss of
influence only because the same instrument, on the same firms, resolves
clearly nonzero effects at shorter lengths. Where the instrument cannot
resolve an effect, we say so, rather than reporting an uninformative zero.

\subsection{Models}

Table~\ref{tab:design} lists the model set. The primary family is Qwen3.5:
a 9-billion-parameter base model serves as the workhorse for all
internal-representation analysis, and post-trained 2B, 9B, and 27B variants
form a scale ladder that asks whether capability changes the phenomenon
rather than merely its level. We flag the workhorse choice openly: it is
dictated by instrumentation, not by behavior---the pretrained sparse
autoencoders our mechanism analysis requires exist only for base
checkpoints---and the base model is not what anyone deploys. The behavioral
results therefore report both 9B variants side by side
(Table~\ref{tab:scale}), including a post-training anomaly discussed in
Section~\ref{sec:phenomenon} that the base model does not share. Because
scale-ladder members must be able to
respond to risk at all for their length profiles to be interpretable, each
model passes a pre-specified capability gate---it must order the seven-level
severity ladder at short context; one candidate control model (a
full-attention model from the prior Qwen generation) failed this gate and is
excluded from length-decay interpretation (Appendix~\ref{app:robustness}).

Two cross-family rows, Llama-3.1-8B and gemma-4-12B, establish that the
retrieval--integration gap is not specific to one model family, at the cost of minor
measurement accommodations documented in Appendix~\ref{app:measurement}.
External validity on current commercial systems comes from qwen3.8-max, a
production API model, with a dated model snapshot retained for
reproducibility, and from Gemini 3.1 Flash-Lite, a low-cost commercial
system measured behavior-only as a capability endpoint
(Section~\ref{sec:phenomenon}; Appendix Table~\ref{tab:app_frontier}). All open-weight inference runs on pinned hardware,
software, and numerical configuration; the same disclosure, at the same
length, always meets the same machine.

\section{The Retrieval--Integration Gap Under a Fixed Information Set}
\label{sec:phenomenon}

This section establishes the retrieval--integration gap empirically. We
first document that the marginal decision influence of the target
disclosure declines sharply as context grows, holding the focal firm's
information set fixed (H1). We then show that the decline is not a failure
of retrieval: under direct questioning, the same AI analysts recover the
disclosure with undiminished accuracy at every length we study (H2), and
the same retrieval--integration gap appears when the disclosures are ones
firms actually filed. We close the section with three results that
discipline interpretation---the effect of model capability, the separation
between ordinal and cardinal sensitivity to disclosure severity, and a set
of pre-specified secondary analyses.

\subsection{Declining Decision Influence as Context Grows}
\label{sec:h1}

\begin{figure}[p]
\centering
\includegraphics[width=\textwidth]{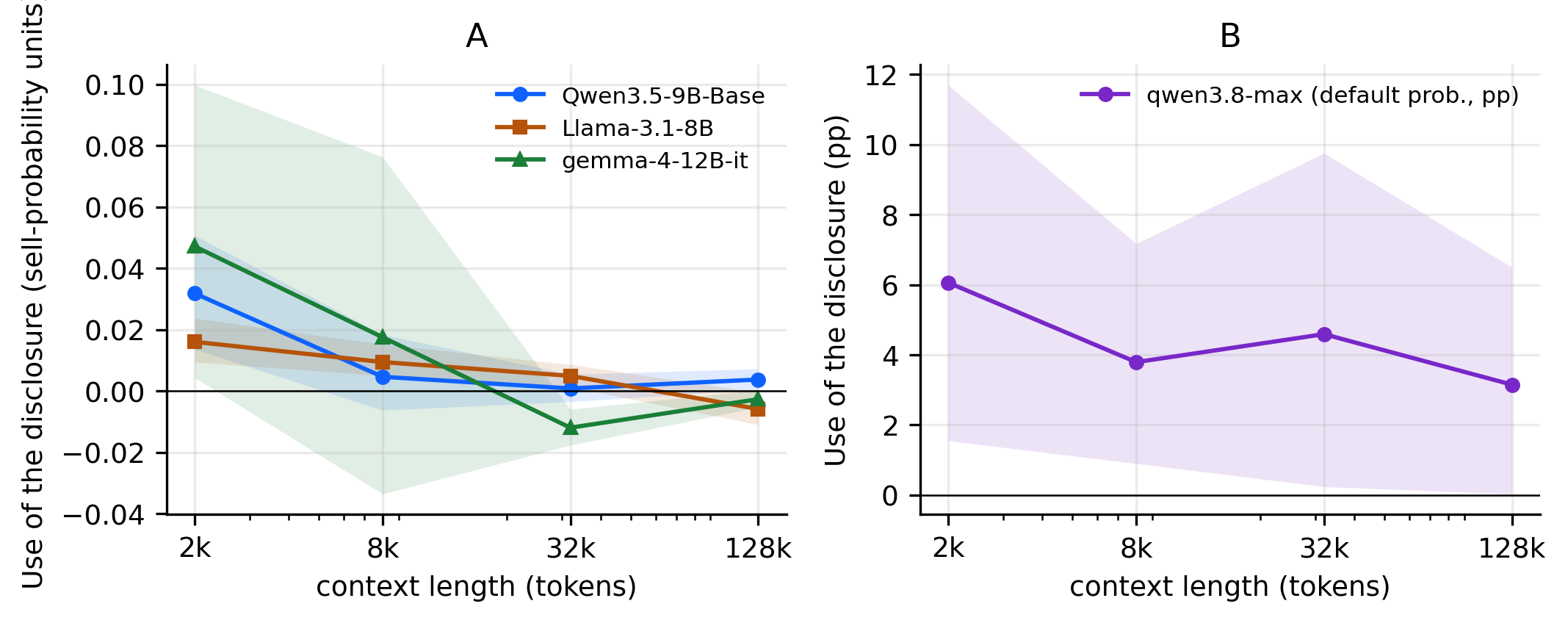}
\caption{\textbf{The marginal decision influence of a fixed disclosure
declines with context length.} Panel A: three open-weight model families,
sell-probability readout, with firm-bootstrap 95\% confidence bands. Panel
B: production API model, unit-bearing numeric judgment. The focal firm's
information set is identical at every length; only economically unrelated
filler text is added.}
\label{fig:h1}
\end{figure}

Table~\ref{tab:main_use} and Figure~\ref{fig:h1} report the main result.
For the 9B workhorse model, the disclosure moves the investment judgment by
$+0.032$ in sell-probability units at 2{,}000 tokens, an economically
meaningful revision given that the neutral-control baseline places
substantial probability on a hold recommendation. As unrelated filler text
expands the context---with the focal firm's information set unchanged---the
influence falls to $+0.005$ at 8{,}000 tokens, $+0.001$ at 32{,}000, and
$+0.004$ at 128{,}000 tokens. Firm-level bootstrap intervals exclude zero
at 2{,}000 tokens in both arrangements ($+0.032$ with filler before the
focal filing, $+0.025$ with filler after) and only sporadically thereafter.

Because a small mechanical effect of inserting \emph{any} paragraph could
masquerade as ``use,'' our inference compares the target
effect not with zero but with an empirical null distribution: the same
insertion machinery applied to ten irrelevant pseudo-disclosures per firm
(Appendix Table~\ref{tab:app_permutation}). Against this null, the 9B
model's use is resolvable at 2{,}000 tokens (permutation $p=0.025$ and
$p=0.0015$ in the two arrangements) and at 8{,}000 tokens in the
after-arrangement ($p=0.038$); at 32{,}000 and 128{,}000 tokens the target
effect is statistically indistinguishable from an irrelevant insertion (all
$p>0.5$). This structure is what makes the long-context null
interpretable: the same instrument that resolves non-zero use at short
lengths finds none at long lengths, so the 128k result reflects a genuine
loss of influence rather than a loss of measurement resolution.

Panel B of Table~\ref{tab:main_use} repeats the design on a current
production model accessed through its commercial interface, using a
unit-bearing readout: the model's stated twelve-month default probability.
The disclosure raises the stated default probability by $6.06$ percentage
points at 2{,}000 tokens, and by $3.79$, $4.59$, and $3.15$ percentage
points at 8{,}000, 32{,}000, and 128{,}000 tokens. Two complementary tests
apply. Against zero, the effect remains significant at every length,
including 128k. Against the empirical insertion null, the effect is
resolvable through 32{,}000 tokens ($p=0.0005$) but falls inside the
insertion-noise band at 128{,}000 tokens ($p=0.11$). The two tests answer
different questions, and we report both: at 128k the production model's
response to the disclosure is nonzero, yet no larger than its response to
an economically irrelevant paragraph subjected to the same insertion.

The pattern replicates across model families (Figure~\ref{fig:h1}, Panel
A). In the Llama family, use declines monotonically from $+0.016$ at
2{,}000 tokens to $-0.006$ at 128{,}000; in the Gemma family, from
$+0.047$---the largest short-context response among the open-weight
models---to $-0.012$ at 32{,}000 tokens, where only one of twelve firms
retains a positive effect. Three architectures with different attention
designs, training data, and providers produce the same qualitative
phenomenon: strong use of the disclosure in short contexts, decay toward
(and occasionally through) zero as unrelated text accumulates. The decay
is therefore not an idiosyncrasy of one model family, and---because the
focal firm's information is fixed by construction---it cannot reflect
rational updating on new information. The design's primary contrast can
also be tested directly as a within-firm paired difference,
$\mathrm{Use}_{2k}-\mathrm{Use}_{128k}$: the decline is $+0.026$ for the
workhorse model with arrangements pooled (exact sign-flip $p=0.0044$),
$+0.015$ for Llama ($p=0.0010$), and $+0.040$ for Gemma ($p=0.0352$),
while the production model's paired decline ($+2.47$ percentage points)
is not statistically resolved ($p=0.199$) and its curve should be read
as a point-estimate pattern (Appendix
Table~\ref{tab:app_paired_decline}). Related work documents that input
length degrades task accuracy \citep{levy2024same, liu2024lost}, and does
so even when retrieval is verified perfect
\citep{du2025context}; our contribution here is
to show the analogous result for the economic weight a model places on a
financial disclosure, under a design that separates pure context load from
information content.

Nor does the decline depend on how the judgment is elicited. Asked
instead whether a credit-rating downgrade is likely---a different
economic question, read out through the same one-token machinery---the
workhorse model's use of the disclosure falls monotonically from
$+0.049$ at 2{,}000 tokens, an effect roughly seven times the dispersion
across neutral versions, to $+0.006$ at 128{,}000 in the
after-arrangement (Appendix Table~\ref{tab:app_second_dv}). The decline
is a property of the judgment stage, not of one question's phrasing.

\subsection{What the Disclosure Still Does: Ordinal Without Cardinal}
\label{sec:severity}

The single-disclosure effect of the previous subsection has a natural
yardstick: how much can this disclosure move the judgment at all? To
answer, each firm's target exists in seven calibrated severity variants,
from nearly innocuous to near-default, and the ladder is run in full at
both endpoint lengths. At 2{,}000 tokens the primary model's judgment
tracks severity clearly in aggregate: the most severe variant draws a
higher sell probability than the mildest for twelve of twelve firms,
severity-level means pooled across firms rise monotonically through the
ladder, and pairwise ordering accuracy is 0.929---though the full
seven-level ordering is strictly monotone within only two individual
firms. The pooled response spans 0.23 in sell-probability units from the
mildest to the most severe variant---the scale against which the
baseline effect of $+0.032$ (roughly 14\% of the full span) should be
read. At 128{,}000 tokens two things happen at once
(Figure~\ref{fig:severity}): the \emph{range} of the response collapses
from 0.23 to 0.04---a factor of 5.6---while the model's ability to
\emph{order} severities degrades only modestly (pairwise ordering
accuracy falls from 0.93 to 0.79). Because the
compression statistic is a ratio of the model's own response spans, it is
unit-free and does not depend on the noise-floor argument that bounds the
single-disclosure result. The production model shows the same asymmetry
in attenuated form, retaining 88\% of its severity response range at
128k (35.8 to 31.5 percentage points).

The disclosure's surviving footprint at long context is therefore ordinal
rather than cardinal: the model still partially ranks worse news as worse,
but transmits almost none of the magnitude into its judgment. An investor
relying on such a system would receive directionally sensible but
economically flattened judgments---a risk ranked correctly but weighted at
close to nothing.

\subsection{The Decline Is Not Retrieval Failure}
\label{sec:h2}

A natural reading of Section~\ref{sec:h1} is that the model simply loses
the disclosure in the haystack: with 128{,}000 tokens in context, it can no
longer find the covenant threshold or the settlement amount, so the
disclosure cannot affect its judgment. Table~\ref{tab:recognition} and
Figure~\ref{fig:reading_using} reject this reading.

In a separate call on the identical documents, we ask each model directly
what the filing says about the target matter and score its answer against
frozen fact keys. The 9B model passes this retrieval test for twelve of
twelve firms at every context length, including 128{,}000 tokens, where its
behavioral use of the same facts is statistically indistinguishable from an
irrelevant insertion. The Gemma model retrieves the disclosure for twelve
of twelve firms at 2{,}000 tokens and ten of ten at 128{,}000 tokens---at
which lengths its measured use is approximately zero---and the production
model's retrieval completeness is essentially flat (0.77, 0.76, 0.72, 0.74)
across the four lengths. Neutral-control filings discipline these numbers:
the Qwen models produce zero false-positive ``retrievals'' on filings from
which the target was removed, across hundreds of control calls, so the
retrieval scores do not reflect a model willing to assert facts that are
not in the document.\footnote{The Llama and Gemma models do show 5--25\%
false-positive rates on neutral filings---they occasionally claim to have
found risk disclosures that are not present. Raw pass rates therefore
overstate those families' retrieval, and their retrieval evidence should
be read through the target-minus-neutral retrieval contrast, which
remains large for every family.}

The divergence between these two measurements is the
retrieval--integration gap, and it is the central fact of the paper.
At 128{,}000 tokens the model can state the covenant threshold, the
remaining headroom, and the consequence of breach---and its investment
judgment does not move. In the language of disclosure processing costs
\citep{blankespoor2020disclosure}, our design holds the awareness stage
fixed---the retrieval question names the issue---and shows that,
conditional on that naming, directed retrieval stays essentially perfect
at every length we study. What does not survive is the integration of
the located disclosure into a revised judgment. Within the stages the
design can separate, integration, not retrieval, binds. This is also the
evaluation point: retrieval accuracy is an observable and convenient proxy
for whether an AI analyst has processed a disclosure, but it does not
establish decision usefulness. An artifact audited only on what it can
state would pass every check in Table~\ref{tab:recognition} while its
investment judgments remained invariant to the facts it demonstrably
retrieved.

The gap is not an artifact of our constructed mini-filings. In an
ecological validation, we surgically insert the target disclosure into the
firms' complete, investor-visible 10-K filings (removing any conflicting
original passages through recorded, machine-verified edits). On these
complete filings, retrieval completeness is 0.931 while measured use is
approximately zero (Appendix Table~\ref{tab:app_ecological})---the same
retrieval--integration gap, in the information environment investors
actually face.

\subsection{Real Disclosures, Surgically Redacted}
\label{sec:realdisclosures}

Constructed materials buy identification; the natural objection is that
investors do not read researcher-authored paragraphs. The redaction
design of Section~\ref{sec:design} addresses it on twenty actual
filings, using disclosures the registrants themselves wrote, in an
analysis we label exploratory and capability-conditional. On the most
capable open-weight model, removing a real disclosure from a roughly
2{,}000-token excerpt moves the investment judgment by $+0.037$ on
average across all twenty filings, with a confidence interval that
narrowly includes zero ($[-0.003, +0.079]$); on the sixteen disclosures
labeled unambiguously adverse, the effect is $+0.042$
with an interval excluding zero ($[+0.003, +0.089]$). The same removal
from the complete filing moves the judgment by $+0.000$
($[-0.003, +0.003]$), indistinguishable from the placebo edit
($+0.001$), while directed retrieval on the complete filings passes for
twenty of twenty firms (Table~\ref{tab:real_disclosure}). A disclosure
the model demonstrably reads---written by the registrant, in the place
the registrant put it---appears to influence the judgment when the
surrounding document is short and loses that influence entirely at full
length.

Two qualifications attach. First, the workhorse 9B model shows no
excerpt-level response to real disclosures ($-0.006$). Real filings are
denser and less curated than the constructed materials, and a mid-sized
model does not extract a usable signal from a single real paragraph
even at short length; the real-disclosure result therefore rests on the
capable model, a capability moderation consistent with the next
subsection. Second, real disclosures are directionally
heterogeneous---a settlement can resolve a litigation overhang, and a
covenant paragraph can certify comfortable compliance---which is why the
adverse-only subset is the sharper test, and why the constructed
materials, with their calibrated severity, remain the workhorse of the
design.

\subsection{Capability Moves the Horizon Outward}
\label{sec:capability}

Does the problem disappear as models improve? Table~\ref{tab:scale} and
Figure~\ref{fig:scale} address the question within a single model family,
holding tokenizer lineage and training pipeline approximately constant.
The smallest (2B) model shows weak, noisy use---it responds to the
disclosure at 2{,}000 tokens ($+0.023$) but discriminates within a
compressed band around a strong prior toward selling, and its length
profile is not monotone (Appendix Table~\ref{tab:app_sevgate} reports the
capability gate that documents this compression). The 9B base model
resolves use at 2{,}000--8{,}000 tokens, as above. The 27B model uses the
disclosure far more at every length ($+0.105$ at 2{,}000 tokens) and is
the only open-weight model whose use remains statistically significant at
every length in both arrangements, with a 128k effect ($+0.034$) roughly
equal to the 9B models' short-context effect. The production-scale
commercial model, as noted, resolves use through 32{,}000 tokens.

A low-cost commercial system from a third family anchors the bottom of
the capability range (Appendix Table~\ref{tab:app_frontier}). With its
reasoning fixed at the lowest setting its provider allows, Gemini~3.1
Flash-Lite shows the retrieval--integration gap on the recommendation
readout: its use of the disclosure falls from $+0.165$ at 2{,}000 tokens to $+0.042$
at 32{,}000 and $+0.029$ at 128{,}000---retaining roughly 18 percent of
its short-context level---and its severity span compresses by a factor
of 2.5 (45.0 to 18.0 percentage points), between the open-weight
workhorse's 5.6 and the production system's 1.14. Uniquely among the
systems we study, its \emph{retrieval} also degrades, from twelve of
twelve firms at 2{,}000 tokens to seven of twelve at 32{,}000 and
128{,}000. Even there the gap retains its shape---retrieval falls to
58 percent of its short-context level while use falls to 18
percent---but the result marks a capability floor below which the
lookup stage itself begins to fail. Capability, in short, buys distance
from the failure at both ends of the commercial range, and several
currently deployable open-weight and commercial systems---every
open-weight model in our ladder and the production system
above---operate well inside their failure horizons.

Table~\ref{tab:scale} also reports the post-trained 9B variant, and we
flag its behavior openly because it is the one result in the family that
does not fit the pattern. In the before-arrangement shown in the table,
the post-trained 9B responds to the risk disclosure with a significantly
\emph{negative} use at 8{,}000 tokens and beyond ($-0.030$ to $-0.051$,
all intervals excluding zero), while in the after-arrangement its use is
significantly positive at the same lengths ($+0.009$ to $+0.019$;
Appendix Table~\ref{tab:app_inversions}). This is not an artifact of
low capability---the post-trained 9B has the \emph{strongest} severity
discrimination in the entire ladder ($S_7-S_1$ range of $+0.42$, versus
$+0.23$ for its base counterpart; Appendix
Table~\ref{tab:app_sevgate})---and we do not have a mechanism for it.
What the pattern indicates is that post-training changes not only how
strongly a model responds to a risk disclosure but the sign structure of
that response across document layouts, an instability that the base model
does not share and that deployment-oriented evaluations of AI analysts
should be aware of. The capability conclusion above does not depend on
this rung: it rests on the 2B--27B endpoints and on the production model,
which agree.

We summarize this pattern as an \emph{integration horizon that moves outward
with capability}: more capable models sustain the disclosure's influence
deeper into the context. Within every model whose
horizon lies inside our design, the influence at the longest contexts is
a small fraction of its short-context value: the horizon moves, but
nothing in the ladder suggests the gap closes except by buying a more
capable reader. For practice, the implication is double-edged.
Disclosure effectiveness under AI readership depends on which reader an
investor can afford, and the context lengths at which mid-tier
commercial systems are marketed for full-filing analysis are precisely
the lengths at which those systems no longer distinguish the target
disclosure from an irrelevant insertion.

\subsection{Secondary and Additional Analyses}
\label{sec:secondary}

\emph{Placement and distance of the added text.} Our design varies whether
the filler precedes or follows the focal filing, a contrast pre-specified
to speak to retention-based mechanisms. Once use itself is small, however,
this difference-of-differences asks more resolution of the behavioral
instrument than it has: on the workhorse model, at every length, the
placement gap is within the range produced by the insertion machinery
alone (Appendix Table~\ref{tab:app_h3}); on the post-trained 9B the
contrast is confounded by the sign instability just discussed. The
fixed-total factorial of Section~\ref{sec:design} speaks to one side of
the question directly: holding total context at 32{,}000 tokens while
the disclosure's distance to the decision question moves from 1{,}000
to 24{,}000 tokens---or at 128{,}000 tokens while it moves from
1{,}000 to 64{,}000---leaves use flat at the noise floor in every cell
(Appendix Table~\ref{tab:app_factorial}). At lengths past the failure
horizon, then, proximity does not rescue the disclosure: bringing it
within 1{,}000 tokens of the decision recovers nothing. Below the
horizon, where use is alive, two direct contrasts speak to the
explanations. Holding total context at 8{,}000 tokens and moving the
disclosure from 1{,}000 to 5{,}500 tokens before the decision changes
use by less than $0.0001$ in absolute value (95\% CI $[-0.007,+0.009]$,
exact $p=0.99$); holding
distance at 1{,}000 tokens and raising total context from 8{,}000 to
32{,}000 lowers use by $0.008$ (CI $[+0.002,+0.015]$, $p=0.03$). In the
tested cells, then, total context load matters and a moderate change in
distance does not---evidence supporting, though not by itself fully
separating, the two explanations, and consistent with the
capacity-limited running summary Section~\ref{sec:mechanism}
examines.

\emph{Repetition.} Disclosure salience matters when the reader's attention
is limited \citep{hirshleifer2003limited}, and repeating a disclosure is a
natural prescription for raising it. Repeating
the target paragraph (holding total length and the position of the final
occurrence fixed) helps only partially and not uniformly: four
occurrences raise use above the single-occurrence level in the pooled
length cells, while two occurrences do not improve every cell, and at
128{,}000 tokens even four occurrences deliver less than a third of the
influence a single occurrence commands at 2{,}000 tokens. Repetition
partially refreshes, but does not restore, the disclosure's weight.

\emph{Semantically similar context.} When the added text consists of other
firms' risk disclosures of the same type---rather than economically
unrelated material---the production model's use falls further ($+3.7$
versus $+5.5$ percentage points at 2{,}000 tokens) and turns negative at
8{,}000 tokens ($-4.3$), consistent with interference from competing,
similar-sounding information rather than pure context load
\citep[cf.][]{hirshleifer2009driven}. The local 9B model shows no such
separation, suggesting that susceptibility to semantic interference is
itself capability-dependent.

\emph{Sign inversions.} Finally, in two model families the point estimate
of use turns \emph{negative} at long context with confidence intervals
excluding zero (Gemma at 32{,}000 tokens, $-0.012$; Llama at 128{,}000
tokens, $-0.006$; Appendix Table~\ref{tab:app_inversions}): the risk
disclosure makes these models marginally \emph{less} inclined to sell.
With roughly forty cells tested, about two such stars would be expected by
chance, and we treat the inversions as suggestive. We note them because
they echo across independent architectures at adjacent lengths, and
because a mildly contrarian response to half-processed risk news is, if
real, more troubling for users than simple neglect.

\section{Mechanism: Where the Disclosure Is Lost}
\label{sec:mechanism}

The behavioral evidence in Section~\ref{sec:phenomenon} establishes the
retrieval--integration gap: a disclosure the AI analyst can retrieve on
demand loses its influence on the judgment it produces as context grows.
Behavior alone, however, cannot say where inside the artifact the
disclosure is lost. Three accounts are consistent with the same outward
pattern. The system may never form a
usable internal representation of the disclosure once the document is
long. It may form that representation but fail to carry it forward to the
point at which the judgment is produced. Or it may carry the information
forward and simply fail to weight it. These accounts have different
implications for design and practice---the first is a reading failure, the
second a transmission failure, the third a judgment failure---and they
suggest different remedies. Because the delegated analyst is an artifact
rather than a person, we can do what archival research on human information
processing cannot:
open the processor and measure its intermediate states directly
\citep{blankespoor2020disclosure}.

\subsection{Two Instruments}

We measure internal states through two independent instruments. The first is
a vocabulary-space probe \citep{gurnee2026verbalizable}, which reads any
internal representation out along directions that correspond to the
model's own vocabulary; intuitively, it asks what words an internal state
``says,'' and how loudly, at any point in the document and at any depth in
the model. To our knowledge this is the first application of the method in
information systems research. The second instrument is a dictionary of interpretable
directions estimated by a sparse autoencoder \citep{qwen2026scope,
templeton2024scaling}; from this dictionary we freeze a small set of
risk-concept features---directions that activate on covenant, default, and
litigation language---and measure how strongly they fire at chosen
positions. The two instruments are estimated independently and measure
different things: the probe measures whether particular words are
\emph{readable} from a position; the feature dictionary measures whether
particular \emph{concepts} are present there. Implementation details,
layer selection, and calibration are in Appendix~\ref{app:mechanism_methods}.

Both instruments are applied at two locations in every filing: the
\emph{disclosure position}---the tokens of the target paragraph
itself---and the \emph{decision position}, the final token before the
model produces its recommendation, where whatever information the judgment
draws on must be available.

\subsection{The Disclosure Stays Encoded Where It Was Read}

The first result rules out the reading-failure account. At the disclosure's
own position, the probe finds the risk vocabulary of the target paragraph
readable at essentially constant strength regardless of document length:
the median rank of the best risk word is stable (approximately 80, 52, and
67 at 2k, 32k, and 128k tokens) at the model's reading layer. The feature
dictionary agrees: risk-concept activation at the disclosure position is
flat from 2k to 128k tokens and strongly target-specific at the
encoding layer (roughly sixty times more intense on target filings than
on neutral controls at 2k, and thirty-fold at 32k and 128k). Whatever the model does
with the disclosure downstream, it forms the same local representation of
it in a 128k-token filing as in a 2k-token one. This is the internal
counterpart of the retrieval result in Section~\ref{sec:phenomenon}: the
model reads the disclosure, and its reading does not degrade.

\subsection{The Decision Stage Never Holds the Disclosure---and Degrades
with Length}

\begin{figure}[p]
\centering
\includegraphics[width=\textwidth]{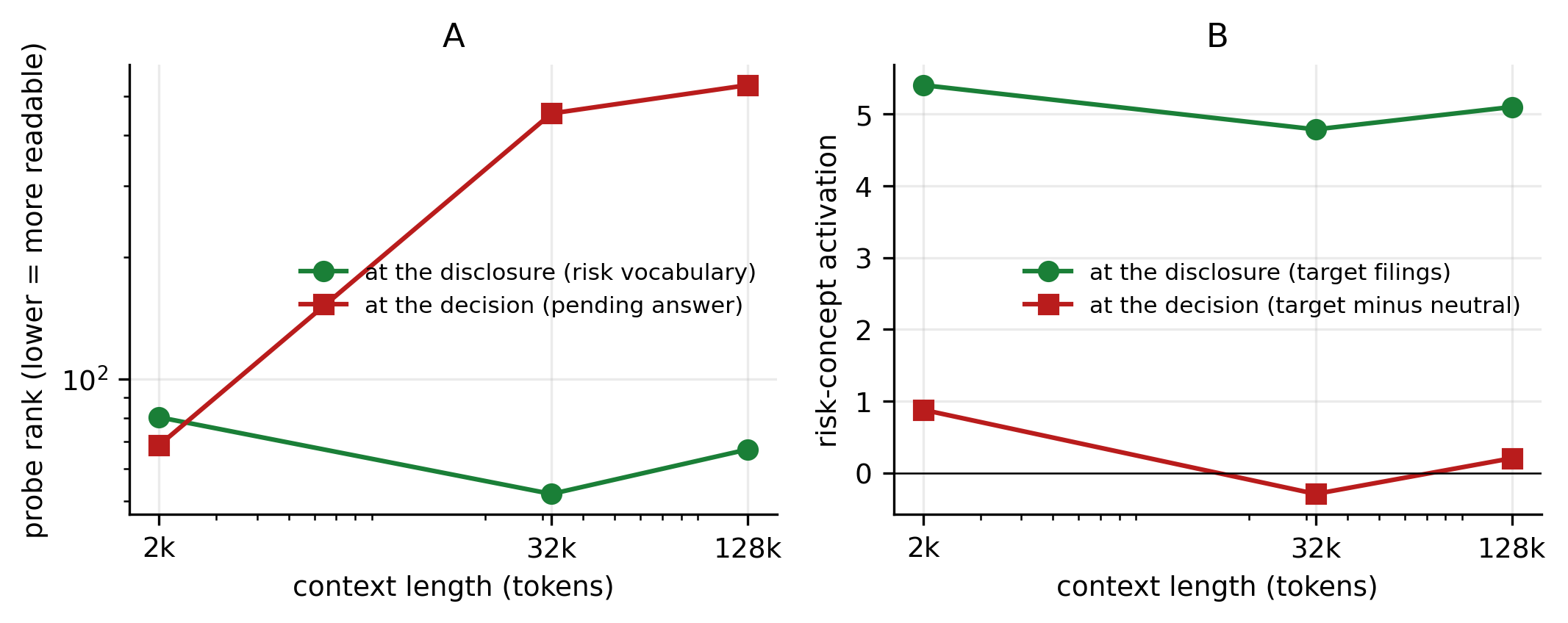}
\caption{\textbf{Where the disclosure is lost.} Panel A: a vocabulary-space
probe shows the risk content of the disclosure remains readable at the
disclosure's own position at every length (green), while the legibility of
the model's pending answer at the decision position degrades roughly
eightfold with length (red); this degradation is common to target and
neutral filings, so it is a property of the decision stage, not of the
disclosure. Panel B: sparse-autoencoder risk features confirm the
disclosure's conceptual encoding is stable at its own position, while
neither instrument detects disclosure-specific content at the decision
position at any length --- including lengths at which behavior responds
strongly. Together the panels indicate that the judgment draws on memory
formed earlier rather than on content held at the decision position, and
that the decision stage's condition degrades with length.}
\label{fig:mechanism}
\end{figure}

The decision position tells a different story, and it is sharper than a
simple mirror image. Neither instrument detects disclosure-specific
signal at the decision position at any length. The feature dictionary
finds no disclosure-specific risk content there: target and neutral
filings are statistically indistinguishable even at 2k tokens, where
behavior responds strongly (Panel~B of Figure~\ref{fig:mechanism}). The
probe agrees: the legibility of the model's pending answer at the
decision position is the same on target and neutral filings (median
target-to-neutral ratio approximately 1.0 at every length). What the
probe adds is that this legibility---whatever the filing
contains---degrades roughly eightfold as context grows (the median rank
of the best decision token moves from 68 at 2k to 454 at 32k and 533 at
128k tokens; Panel~A of Figure~\ref{fig:mechanism}). Both instruments
test particular signal classes---vocabulary-readable directions and a
frozen feature dictionary---so these nulls bound what the decision
position \emph{detectably} holds rather than everything it could
possibly encode.

Read together, these facts favor a particular architecture of the
failure. If the decision position detectably holds no disclosure content
even when influence is strong, a natural account is that the influence is
mediated by the model's ability to draw, at the moment of judgment, on
memory formed earlier---whether by reaching back to the disclosure's own
position, where the content remains fully intact, or through an
intermediate store. A division of labor in which content stays where it
was encoded and is fetched when needed is the account of factual recall
developed in the interpretability literature \citep{geva2023dissecting,
wu2025retrieval, feng2024binding, olsson2022induction,
variengien2023look}, and it gives representational form to the
behavioral finding of \citet{du2025context} that length harms
performance even when retrieval is perfect. On this reading, the failure
documented in Section~\ref{sec:phenomenon} is a \emph{transmission}
failure located at integration rather than at comprehension: the reading
is intact, the reader is intact, and what degrades with length are the
channels connecting them at decision time. The
causal experiments below ask which channels those are.

Two validation results discipline this interpretation. First, our primary
model family interleaves fixed-size linear-attention layers with standard
full-attention layers, and fixed-size recurrent state is a natural suspect
for retention pressure \citep{jelassi2024repeat, arora2024zoology}; the
probe, however, had previously been validated only on standard
architectures. A calibration exercise across adjacent layer pairs shows
that readability tracks depth, not layer type---adjacent linear- and
full-attention layers agree pairwise---so the decay we measure is not an
artifact of the architecture's unusual layers
(Appendix~\ref{app:mechanism_methods}). Second, the decay reproduces
across independently chosen layers and across the two endpoints of the
severity ladder, so it is not a peculiarity of one probe location.

The remainder of this section probes the transmission account causally by
editing the artifact's internal computation directly. The architecture
provides two transmission channels through which information read earlier can reach the
judgment: a fixed-size recurrent memory state that summarizes preceding
content, referred to intuitively below as a \emph{running summary}, and
attention-based lookup over the source text. The verdict, stated in
advance: both channels carry the disclosure. Matched-slot
blackouts produce a disclosure-specific attention effect equal to 44 to
48 percent of the disclosure's contrast; channel transplants show the running
summary removing about two-thirds and recreating nearly half; and the
two estimated channel effects are statistically indistinguishable in
size, so the result is a dual-carrier identification, not a ranking.

\subsection{Which Channel Carries the Disclosure to the Judgment}
\label{sec:mechanism_causal}

A language model of our workhorse's class maintains both transmission
channels while it reads. The first is a lookup store: the attention
mechanism keeps
every token's representation and can, in principle, consult any of them
when a later computation asks for it. The second is the running summary: the
model's linear-attention layers compress everything read so far into a
recurrent state of fixed size, updated token by token, in the manner of a reader
keeping cumulative notes rather than re-consulting the document
\citep{katharopoulos2020transformers, yang2025gated}. A fixed-size summary
has a known limitation: its capacity to retain any individual fact falls
as more text competes for the same space
\citep{arora2024simple, jelassi2024repeat, arora2024zoology}. This
combination of the two channels is a design shared by a growing set of
recently released long-context architectures
\citep{yang2025gated, qwen2026scope}, so the question of which channel
carries a disclosure into a judgment is not a curiosity of one model.

We first locate the lookup route. Measuring, at the moment of decision,
how much attention each head directs at the disclosure's tokens---against
matched neutral text in the same slot---identifies a small set of heads
whose reading is content-driven with high cross-firm consistency (eight
to twelve of twelve firms across the five most selective heads); in the hybrid workhorse these heads concentrate in the
model's later layers, while in the pure-attention comparison architecture
the most selective heads split between the first layer and the late
layers (Appendix Table~\ref{tab:app_attn_locate}). At full length this
content-selective reading collapses to a few percent of its
short-document strength in both architectures while directed retrieval
remains intact: the lookup route visibly exists, and visibly withers
with length.

Visibility, however, is not causal weight. Severing the identified
heads' access to the disclosure---zeroing their attention to its tokens
throughout the decision computation, with random-head and wrong-span
controls, following standard practice for such interventions
\citep{meng2022locating, wang2023interpretability, zhang2024towards}---
leaves the judgment unchanged: the selective heads are individually
redundant. Severing \emph{every} attention head's access is a different
matter. The operative matched control is the identical blackout applied
to the neutral document's slot, and it is cleanly null; netting it out,
the disclosure-specific effect of removing all attention access is 44 to
48 percent of the disclosure's own contrast, in ten to eleven of twelve
firms depending on scope (Appendix Table~\ref{tab:app_attn_intervene}).
A further sham that blacks out a content-free span proved unstable
across firms and is reported as a failed control rather than used in
the identification, which rests on the matched neutral-slot difference.
The lookup route, once properly identified, carries
a substantial share.

A complementary intervention concerns the running summary. Because target and neutral filings are byte-identical outside
the disclosure's slot, both can be processed to the end of that slot and
their two memories recombined---the lookup store from one reading
spliced with the running summary from the other---before the identical
remainder of the document is processed (Table~\ref{tab:transplant}).
When the model reads the target filing but its running summary is the
one formed on the neutral filing, roughly two-thirds of the disclosure's
influence disappears (a ratio of full-sample means of 65 percent)
although the disclosure sits intact in the lookup store. When the model
reads the neutral filing but its running summary is the one formed on
the target filing, nearly half of the influence appears from nothing (44
percent) although the disclosure is nowhere in the document being
processed. A neutral-to-neutral sham transplant---identical surgery,
nothing to transmit---moves the readout by 0.046 in unsigned magnitude,
37 percent of the removal effect and 54 percent of the smaller
recreation effect, bounding the generic component of the intervention.
The sham's signed per-firm mean, however, is $+0.008$ (95\% CI
$[-0.010,+0.024]$): the surgery perturbs the readout without pushing it
in any particular direction, and the real removal exceeds the sham
within firm by $+0.116$ (eleven of twelve firms, exact sign-flip
$p=0.002$; Table~\ref{tab:transplant}).
Both real directions are consistent in ten of twelve firms; one firm
responds to neither transplant and is reported as an exception, and the
remaining limitations---a single neutral counterfactual and no
additivity across channels---are stated in the table and appendix.

The picture these interventions paint jointly is of two working
transmission channels
rather than one: the running summary and attention-based lookup
each carry a substantial share of the disclosure's influence, the two
estimated effects are statistically indistinguishable in size when
compared firm by firm (paired difference of the removal against the
directional attention removal, $+0.025$, exact sign-flip
$p=0.54$; Appendix Table~\ref{tab:app_attn_intervene}), and they need
not, and do not, sum to one, because the channels feed each other during
processing. What
unifies them is what happens under length. The distance experiment
speaks to the two candidate explanations: at below-horizon totals the
disclosure's influence is alive and flat in its distance to the
decision in the tested cells, while it declines significantly in total
context at fixed distance
(Section~\ref{sec:phenomenon}, Appendix Table~\ref{tab:app_factorial})
---a pattern supporting capacity pressure on memory formed during
reading over a lengthening reach at decision time. Directed retrieval
survives because a question that names the disclosure can redirect the
intact lookup route. The response range compresses before the ordering
does, as a fading trace loses magnitude before sign. And the earlier
finding that no single internal \emph{direction} carries the judgment
(Appendix~\ref{app:interventions}) fits an account in which the
operative state is distributed and compressed rather than localized
\citep{geva2023dissecting, vig2020investigating, makelov2024subspace}.

The mechanism conclusion we defend is therefore the following: the
retrieval--integration gap is a failure of integration rather than of
comprehension---the disclosure remains encoded and retrievable, the
judgment machinery is intact, and what weakens as competing context grows
are the channels that carry the disclosure into the judgment---and both
channels, the fixed-size recurrent running summary and attention-based
lookup, are causally implicated as substantial carriers,
with no resolvable ranking between them. The dilution dynamics of the
summary itself are consistent with every behavioral signature but await
designs that manipulate its capacity directly. Locating the failure in
transmission rather than in reading also yields a design implication we
take up next: if decision-relevant information is lost between reading and
judgment, the leverage lies in how a workflow routes that information to
the point of judgment rather than in how much computation the artifact
expends once it arrives there. Section~\ref{sec:remedies} tests
that implication against alternative workflow architectures.

\section{Workflow Architecture: Routing, Not Computation}
\label{sec:remedies}

\begin{figure}[p]
\centering
\includegraphics[width=0.75\textwidth]{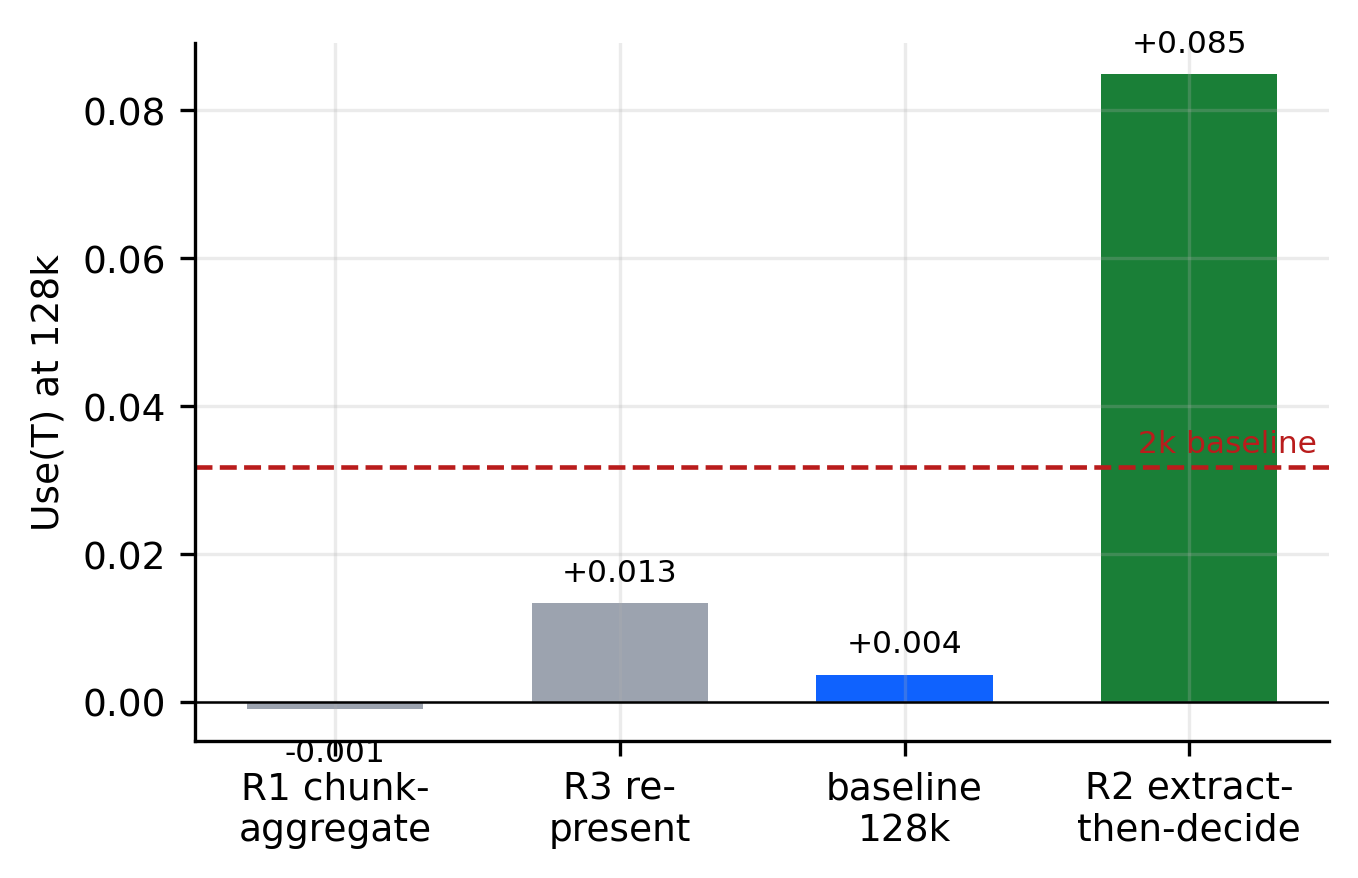}
\caption{\textbf{The remedy gradient at 128k tokens.} Replacing the filing
with the model's own notes destroys the disclosure's influence; verbatim
re-presentation leaves it unchanged; asking the model
to answer the retrieval question first and decide with its own answer in
view raises the share of short-context influence retained at 128k from 12
to 67 percent (dashed line: the unassisted 2k baseline).}
\label{fig:remedies}
\end{figure}

If the mechanism section is right that the disclosure is lost in
transmission---read faithfully, encoded stably, but no longer legible at
the point of judgment---then the retrieval--integration gap should be
closable by rearchitecting the workflow that carries information from
reading to judgment, and closable in a particular way: workflows that
route the disclosure's content back within reach of the
decision should work, while interventions that merely add computation, or
that sever the judgment from the original document, should not. Our fourth
hypothesis (H4) predicted that
reorganized processing---chunked summarization, forced reconciliation of
the retrieval answer with the decision, and pre-decision
re-presentation---would restore the disclosure's influence at long
context. As the results below show, only one of the predicted
reorganizations does, so H4 is partially supported, and the failures of
the other two predicted remedies turn out to be as informative as the
success. The workflow experiments test four protocols,
each an alternative workflow architecture tied to the same outcome
variable
$\mathrm{Use}(\ell)$, on the same twelve firms and the same workhorse
model as the main design (Section~\ref{sec:design}). Table~\ref{tab:remedies}
and Figure~\ref{fig:remedies} summarize the results at 128k tokens;
Appendix Table~\ref{tab:app_remedies_full} reports both filler
arrangements. Two further sets of treatments then sharpen the result: a
matched decomposition that probes which ingredients of the successful
workflow matter (Table~\ref{tab:decomposition}), and a replication
of the key workflows on complete real filings
(Table~\ref{tab:eco_remedies}).

\subsection{Four Workflow Architectures}

\emph{Extract-then-decide} asks the model the directed-retrieval question
first---what does this filing say about the target matter, with amounts
and status---and then prepends the model's own answer to the decision
prompt, with the full filing still in context. It changes no information;
it only makes the model restate what it already knows before deciding.
\emph{Chunk-then-aggregate} processes the filing section by section,
collects the model's extraction notes from each segment, and asks for a
decision on the consolidated notes; this is our bounded, generic
implementation of a chunked, note-based workflow used when documents
exceed comfortable context sizes---a different architecture from
retrieval-augmented generation proper, which retrieves from an external
index \citep{lewis2020retrieval}---and our results speak to this
implementation rather than to every variant deployed in practice. \emph{Re-present} places a
verbatim copy of the target paragraph immediately before the decision
question, adapting the recitation-style mitigation proposed in the
long-context literature \citep{du2025context}. Finally, on the production API model we vary
\emph{extended reasoning} across three effort tiers, holding the document
and prompt fixed---the vendor's own mechanism for spending more
computation on a harder problem.

\subsection{A Sharply Ordered Gradient Across Workflows}

The four workflow architectures produce a gradient far more ordered than
we anticipated. The cleanest statistic is the \emph{within-protocol
retention rate}: the share of a workflow's own short-context influence
that survives at 128k tokens. Under the baseline, 12 percent survives
($+0.032 \rightarrow +0.004$). Under extract-then-decide, 67 percent
survives ($+0.126 \rightarrow +0.085$)---the workflow raises the length
retention of the disclosure's influence by a factor of five, with all
twelve firms responding in the same direction at 128k (the
after-arrangement figures are similar; Appendix
Table~\ref{tab:app_remedies_full}). Because the ratio compares each
workflow with itself, it is immune to the observation that the workflow
also amplifies influence at short context ($+0.126$ at 2k)---making the
model restate a disclosure evidently makes it salient wherever it
sits---and the differenced design keeps even the levels honest, because
the neutral-control filings pass through the identical workflow. As
color rather than argument: the workflow's 128k influence exceeds the
unassisted 2k baseline. Retention of 67 rather than 100 percent also
records, honestly, that rearchitecting the workflow attenuates without
eliminating the underlying erosion.

Chunk-then-aggregate sits at the opposite end of the gradient, and its
result is the most practically consequential in the paper. Deciding on the
model's own consolidated notes eliminates the disclosure's influence
entirely: $-0.001$ at 128k, and---strikingly---$-0.001$ at 2k as well,
where the entire document fits within a single chunk and the ``pipeline''
reduces to summarize-then-decide. Sign counts across firms are
indistinguishable from coin flips.

An audit rerun that archives every note verbatim decomposes this zero,
and the decomposition is unanimous (Appendix
Table~\ref{tab:app_r1_audit}). Scored against the same frozen fact keys
used for directed retrieval, the target disclosure is \emph{absent} from
the consolidated notes in twenty-four of twenty-four firm-arrangement
cells at 2k, and in twenty-one of twenty-four at 128k. The notes reveal
why: under a per-segment token budget typical of practical pipelines, the
model transcribes facts from the head of each segment---revenue,
expenses, cash---until our per-segment budget binds (many notes truncate
mid-sentence), and a single-paragraph risk disclosure loses the
competition for note space before the decision stage ever sees it. The
failure is not decision-stage neglect of extracted facts; it is eviction
at extraction. This also explains, in one stroke, why
extract-then-decide sits at the opposite end of the same gradient: the
two workflows run the same model over the same filing, and differ only
in how they route information---in whether extraction is
\emph{targeted} (a directed question about the
matter at hand) or \emph{budgeted and generic} (summarize everything in
$k$ tokens). Targeted extraction retains the disclosure; budgeted
summarization evicts it. The practical reading is stated at the strength
the evidence supports: our bounded, generic implementation of a
retrieve-summarize-decide pattern reproduced none of the disclosure's
decision influence in our setting, at any length.

The remaining two workflows bracket the mechanism. Re-presenting the
disclosure verbatim before the question---injecting the raw source
paragraph where extract-then-decide injects a structured, targeted
answer---moves essentially nothing ($+0.013$ at
128k, seven of twelve firms positive, within the noise band of the
empirical null); what separates the two workflows, whether authorship or
the form the injected text takes, is the question the decomposition
below is designed to probe.
And in the one setting where we test the computation margin---the
production API model's own extended-reasoning modes, a different system
from the workhorse that runs the workflow treatments---no effort tier
restores any influence at 128k tokens, while at 2k tokens enabling
reasoning \emph{reduces} measured use from $+5.5$ to between $+0.7$ and
$+1.9$ percentage points (arrangement-pooled figures; Appendix
Table~\ref{tab:app_thinking} reports the before-arrangement, $+6.1$ to
$+1.9$/$+2.0$). In the tested vendor's reasoning modes, more thinking
about the same context does not put the disclosure back into the
judgment; at short context it appears instead to dilute it.

\subsection{Decomposing the Successful Workflow}

Extract-then-decide bundles several ingredients: the restatement is
produced by the model itself, it answers a targeted question about the
focal matter, it arrives in a compact structured form, and it sits
immediately adjacent to the decision with the filing still in context. To
separate them, we run five matched conditions that inject text into the
identical position, with the identical framing, before the identical
decision question (Table~\ref{tab:decomposition}). The model's own
targeted extraction yields $+0.113$ at 2k and $+0.055$ at 128k
(after-arrangement). An extraction written by the experimenter from the
disclosure's answer key---same facts, same structured form, same token
budget---yields $+0.151$ and $+0.056$: statistically indistinguishable
from the model's own, and if anything larger at short length. The
identified conclusion is therefore precise but narrow: conditional on
the same structured, targeted form and content, the model's own
authorship is not necessary. A targeted extraction produced by a
different, less capable model recovers about a quarter of the effect at
128k ($+0.013$, with an interval that includes zero)---a deficit
consistent with that model's lower extraction fidelity, though the five
conditions jointly vary targeting, structure, compactness, fidelity, and
wording, so no single mediating ingredient is isolated. The two
unstructured conditions recover little: the model's own words in
verbatim form (asked to copy, not to answer) retain a small but reliable
effect at 128k ($+0.018$), and the re-presented source paragraph
essentially none. What the pattern points to is a \emph{targeted,
structured restatement adjacent to the decision} as the operative
package. For practice the authorship result is suggestive good news: it
implies the workflow need not depend on the model's introspection, and
that a research workflow could supply the focal extraction
externally---though we have tested an experimenter-written answer key,
not analyst workflows in the field.

\subsection{Complete Real Filings}

The workflow ordering survives the move from constructed documents to
complete statutory filings. On the twelve complete, surgically edited
real 10-Ks of Section~\ref{sec:design}, direct reading of the full filing
leaves the disclosure without influence ($-0.004$)---the ecological
replication of the main result---and the two poles of the gradient
reproduce around that zero (Table~\ref{tab:eco_remedies}).
Chunk-then-aggregate cannot destroy what direct reading already fails to
deliver, but its notes audit shows the same eviction documented above
operating on real filings: across seven to fifteen chunks per filing, the
disclosure's facts survive into the consolidated notes in zero of twelve
cases. And extract-then-decide does not merely retain influence here---it
\emph{creates} it where direct reading produces essentially none (the
direct baseline is in fact slightly negative, $-0.004$), reaching
$+0.043$ with eleven of twelve firms positive (95 percent interval
$[+0.026, +0.059]$). On real filings
as on constructed ones, whether a disclosure reaches the recommendation
is a property of the workflow architecture, not of the disclosure.

\subsection{What the Gradient Teaches: The Decision-Proximal
Representation Principle}

Two features of the pattern discipline its interpretation. First, the
successful workflow and the failed ones are separated not by whether the
disclosure's substance is placed before the decision---re-present puts
the source paragraph itself there---but by \emph{the form the routed
information takes and what remains in view}. Influence returns when a targeted,
structured answer to the focal question sits adjacent to the decision
with the source document still present, whether the model or the
experimenter wrote it; it largely does not return when the raw passage
is re-presented, and it disappears altogether when the document is
replaced by notes. This is the shape the mechanism evidence would
predict: if the judgment draws on the model's running summary of the
document, a compact restatement placed at the point of decision
re-enters that summary, where a passive copy of source text and the
diluted trace of the original paragraph do not. Second, in the one
computation margin we test---the production model's reasoning
tiers---effort does not substitute for structure. H4 fared worse than
hypothesized on its breadth and better on its point: of the three
predicted reorganizations, chunked summarization and raw re-presentation
failed, and only targeted extract-then-decide restores influence---so
the hypothesis is partially supported, and what the tested workflows
collectively indicate is that \emph{workflow architecture, not added
computation, governed whether a retrieved disclosure was used in our
setting}.

We state the resulting design knowledge as a \emph{decision-proximal
representation principle}: information critical to a judgment should be
extracted into a targeted, structured representation and placed at the
point of judgment, rather than routed through generic,
capacity-constrained summaries. Each clause of the principle is earned by
a different arm of the gradient. \emph{Targeted}: the extraction must
answer the focal question, because budgeted generic summarization evicts
the disclosure at the extraction stage, before the decision stage ever
sees it. \emph{Structured}: a passive verbatim copy of the source
paragraph, placed in the same position and read by the same model,
recovers little, so it is the form the representation takes and not
proximity alone that carries influence. \emph{Decision-proximal, without
displacing the source}: the restatement works while the filing remains in
context, whereas replacing the document with notes removes the
disclosure's influence altogether. The principle is a claim about routing
rather than about computation or about the model itself: the same model,
the same filing, and the same decision question yield materially
different judgments depending only on where, and in what form,
decision-relevant information is placed. Our design supports the
principle as a package---the five decomposition conditions jointly vary
targeting, structure, compactness, fidelity, and wording---and identifies
model self-authorship as unnecessary conditional on that package rather
than isolating which single ingredient mediates the effect. Within the
single vendor whose reasoning tiers we are able to test, added
computation was not a substitute for this reorganization; we do not claim
that computation never helps.

For practice, the gradient cuts in two directions. For the designers of
AI-assisted investment research, it implies that workflow architecture is
a first-order determinant of whether disclosures reach judgments:
mandatory targeted extraction before the recommendation, produced by the
model or, our decomposition suggests, supplied externally, is cheap and
effective, while a generic chunk-and-summarize pattern of the kind we
implement silently discards decision-relevant content even for short
documents, on constructed and real filings alike. For evaluators of
disclosure effectiveness---including regulators asking whether machine
readers ``see'' a required disclosure---it implies that testing retrieval
is not enough. The retrieval--integration gap is a property of the
deployed workflow as much as of the model: a system can pass any
question-answering audit of a filing and still assign the audited
disclosure no weight, so the audit must reach the judgment stage,
because reading is not using.

\section{Conclusion}\label{sec:conclusion}

A model that can recite a disclosure has not necessarily incorporated
that disclosure into its judgment. That is the central fact this paper
establishes, and it survives every robustness lens we apply. Holding a
firm's information set fixed and lengthening only the economically
unrelated surrounding context from 2,000 to 128,000 tokens, the marginal
decision influence of an authored risk disclosure falls from 3.2
percentage points to---and in two model families slightly through---the
design's noise floor. The decline is not a uniform flattening of
judgment: across a seven-step severity ladder the model's response range
compresses by a factor of 5.6 while its ability to order the same risks
degrades far less, so at full length the model can still rank risks but
no longer scales its judgment to them. Directed retrieval of the same
disclosure, meanwhile, stays flat in length. The divergence between what
an AI analyst can state and what moves its recommendation---the
retrieval--integration gap---is therefore a property of the integration
stage rather than of comprehension or recall.

Causal experiments on the model's memory locate that stage concretely.
Transplanting the model's fixed-size recurrent memory state---the
compressed running summary its linear-attention layers maintain over the
document---between matched filings removes roughly two-thirds of the
disclosure's influence when only that state forgets it, and recreates
nearly half of it when only that state retains it. Severing every
attention head's access to the disclosure's text, with matched controls
in which the identical blackout on a neutral document's slot does
nothing, removes 44 to 48 percent of the disclosure's contrast. Both
memories therefore carry the disclosure: the two estimated channel
effects are statistically indistinguishable in size, neither is
exclusive, and we do not claim a ranking between them or an additive
decomposition of the total effect. A running summary has fixed capacity
that must dilute as total context grows, and that account is consistent
with what we observe---influence eroding with total length, proximity to
the decision failing to rescue it past the horizon, direct questioning
still succeeding, and a targeted extract next to the decision restoring
what was lost---though the dilution dynamics themselves await designs
that manipulate them directly. Consistent with that account, the failure
yields to process design rather than to computation. The tested vendor's extended-reasoning
modes do not restore the lost influence; chunk-based summarization
destroys it outright; but requiring the model to answer the retrieval
question first and decide with its own answer in view raises the share
of short-context influence retained at full length from 12 to 67
percent, an influence of 8.5 percentage points at 128,000 tokens.

These results extend information systems research on delegation from the
allocation of work between humans and AI to the execution of work inside
an agentic artifact. Once an end-to-end analytical task has been
delegated, the artifact must carry decision-relevant information across
its own processing stages, and our experiments show that this internal
handoff can fail while every externally verifiable intermediate operation
succeeds. The gap we document is a post-delegation failure in the strict
sense: the agent demonstrably retrieves the fact it was given, and
demonstrably does not use it in the judgment it was delegated to form.
Success on a checkable subtask is thus not evidence of completion of the
delegated objective---a distinction that grows more consequential as
agentic systems execute multistage work with little human visibility into
the stages in between.

The finding also carries a task-aligned lesson for how generative AI
systems are evaluated. Retrieval accuracy is convenient and observable,
but it is a proxy for the decision the system exists to support, not a
measure of it; our evidence shows that retrieval-based evaluations can
certify systems whose judgments are effectively invariant to the facts
they successfully retrieved. Marginal decision influence supplies the
missing outcome measure. It is counterfactual, continuous, and
manipulable, and because the processor can be probed and re-run at will,
it converts questions that are infeasible with human subjects---how
influence varies with placement, repetition, severity calibration, or
surrounding text, holding everything else fixed---into controlled
experiments. Both the measurand and the fixed-information length design
that isolates it are portable to any AI system whose value depends on
whether retrieved information reaches a decision.

Moving from detection to explanation yields design knowledge. The
workflow gradient is sharply ordered and, once the transmission channels
are identified, interpretable: what separates the protocol that works
from the ones that do not is the form the injected information takes and
what remains in view, not merely whether the disclosure's substance sits
near the decision. The design knowledge that follows is the
\emph{decision-proximal representation principle} of
Section~\ref{sec:remedies}: information critical to a judgment belongs in
a targeted, structured representation at the point of judgment rather
than in a generic, capacity-constrained summary. Pipelines that summarize
documents in chunks and reason over the notes---at least in the bounded,
generic form we implement---silently strip disclosures of their decision
influence, and a verbatim audit of the notes shows why: budgeted, generic
summaries evict a single-paragraph disclosure before the decision stage
ever sees it, in twenty-four of twenty-four cases even when the whole
document fits in one chunk, and in all twelve complete real filings we
test. Pipelines that surface a targeted extraction adjacent to the
decision retain it, and since an experimenter-written extraction works as
well as the model's own, the decomposition suggests the fix could deploy
as an externally supplied restatement of the focal disclosure. Model
capability and workflow architecture accordingly play distinct and
complementary roles: capability moves the integration horizon outward, while
workflow architecture determines whether decision-relevant information
reaches the judgment at all.

For financial disclosure processing and machine readership, the results
distinguish machine \emph{readability} from machine \emph{decision
usefulness}. A natural reading of the automation of disclosure processing
is that concerns about document length and complexity lose force once the
marginal reader is a machine; firms already write with algorithmic
readers in mind \citep{cao2023talk}. Our evidence points the other way.
Generative AI does not eliminate disclosure processing costs so much as
relocate them from finding information to integrating it: the integration
stage of a machine reader degrades with length faster than its retrieval
stage---which in our more capable systems does not measurably degrade at
all---so the effective weight of any single disclosure falls as filings
grow, and falls quietly, because retrieval-based checks will not detect
it. Investors and intermediaries should therefore validate an AI analyst
by whether economically meaningful changes in its information produce
appropriate changes in its judgment, and evaluations of disclosure
effectiveness for machine readers should be conducted at the judgment
stage, where the retrieval--integration gap lives, rather than at the
retrieval stage, where it is invisible.

The limitations are those of a first controlled study. The causal
identification of the transmission channels comes from one model family,
whose hybrid design makes the recurrent memory state separable from the
attention store; the behavioral dissociation replicates in two other
families, and the collapse of decision-time reading replicates in a
pure-attention architecture, but how the disclosure travels in
architectures without a recurrent channel remains to be mapped. Our
primary treatments are researcher-authored, a choice that buys severity
calibration and clean identification; the concern that this manufactures
the phenomenon is answered by twenty real disclosures surgically redacted
from actual filings, where the capable model shows the same
gap---influence in a short excerpt, none in the full filing, retrieval
intact throughout---though the mid-sized workhorse does not respond to
real single disclosures even at short length, so the real-filing evidence
is an exploratory, capability-conditional replication rather than a
universal one. Finally, the integration horizon itself is
capability-dependent in both directions: more capable models within a
family sustain influence deeper into the context, while on a low-cost
commercial system even the retrieval stage begins to degrade at
length---a capability floor beneath the gap. Whether a disclosure is
decision useful under AI readership therefore depends on \emph{which}
machine reads the filing, and evaluations of that usefulness cannot
assume any particular reader. Whether future systems close the
retrieval--integration gap at every length at which disclosure is
written, or merely move it, is a question this paper's instrument is
built to keep answering.

\singlespacing
\bibliography{references}

\clearpage

\clearpage
\begin{table}[p]
\centering
\begin{threeparttable}
\caption{Experimental design and sample}
\label{tab:design}
\small
\begin{tabular*}{\textwidth}{@{\extracolsep{\fill}}lp{11.6cm}}
\toprule
Component & Description \\
\midrule
Sample & 12 U.S. registrants' annual reports (10-K filings), screened so that
a single researcher-authored risk paragraph can be inserted without requiring
changes elsewhere in the filing; every paragraph passes a whole-filing
coherence audit. \\
Target ($T$) & A firm-specific risk disclosure with verifiable amounts
(covenant thresholds, settlement amounts, indemnification caps), authored to
be locally insertable in the notes or MD\&A. \\
Neutral controls ($N_1..N_5$) & Five equal-length neutral replacements of the
target paragraph; the marginal decision influence of the disclosure is
$\mathrm{Use}(\ell)=\mathrm{Decision}(T,\ell)-\overline{\mathrm{Decision}(N,\ell)}$. \\
Context length $\ell$ & 2{,}000 / 8{,}000 / 32{,}000 / 128{,}000 tokens; the
focal firm's information set is held fixed and only economically unrelated,
genre-matched filler text is added. \\
Arrangements & Filler placed entirely before or entirely after the focal
mini-filing (pre-specified secondary comparison). \\
Empirical null & Ten pseudo-target insertions ($P_1..P_{10}$) per cell apply
the identical insertion machinery to irrelevant paragraphs, providing the
reference distribution against which target effects are judged. \\
Decision readouts & Deterministic sell-vs-buy readout at a fixed decision
prompt (open-weight models); a unit-bearing numeric judgment (twelve-month
default probability) for commercial API models. \\
Retrieval readout & A separate directed-questioning call scores retrieval of
the target's amounts and status against frozen answer keys. \\
Models & Qwen3.5 family (2B, 9B, 9B-Base, 27B), Qwen3-8B-Base
(full-attention control), Llama-3.1-8B-Instruct and gemma-4-12B-it
(cross-family rows), qwen3.8-max and a dated qwen3.5-plus snapshot (API),
Gemini 3.1 Flash-Lite (behavior-only commercial arm). \\
\bottomrule
\end{tabular*}
\end{threeparttable}
\end{table}

\clearpage
\begin{table}[p]
\centering
\begin{threeparttable}
\caption{The marginal decision influence of a disclosure declines with context length}
\label{tab:main_use}
\small
\begin{tabular*}{\textwidth}{@{\extracolsep{\fill}}lcccc}
\toprule
 & \multicolumn{4}{c}{Context length} \\
\cmidrule(lr){2-5}
Model & 2k & 8k & 32k & 128k \\
\midrule
\multicolumn{5}{l}{\emph{Panel A: open-weight models, }$\mathrm{Use}(\ell)$\emph{ in sell-probability units (filler before target)}} \\
\addlinespace
Qwen3.5-9B-Base & $+0.032^{*\dagger}$ & $+0.005$ & $+0.001$ & $+0.004^{*}$ \\
 & {\scriptsize$[+0.014,+0.051]$} & {\scriptsize$[-0.006,+0.018]$} & {\scriptsize$[-0.004,+0.005]$} & {\scriptsize$[+0.000,+0.007]$} \\
Llama-3.1-8B & $+0.016^{*}$ & $+0.009^{*}$ & $+0.005^{*}$ & $-0.006^{*}$ \\
 & {\scriptsize$[+0.009,+0.024]$} & {\scriptsize$[+0.005,+0.015]$} & {\scriptsize$[+0.001,+0.009]$} & {\scriptsize$[-0.011,-0.001]$} \\
gemma-4-12B-it & $+0.047^{*}$ & $+0.018$ & $-0.012^{*}$ & $-0.003$ \\
 & {\scriptsize$[+0.004,+0.100]$} & {\scriptsize$[-0.034,+0.076]$} & {\scriptsize$[-0.018,-0.006]$} & {\scriptsize$[-0.005,+0.000]$} \\
\addlinespace
\multicolumn{5}{l}{\emph{Panel B: production API model, numeric judgment (percentage points of default probability)}} \\
\addlinespace
qwen3.8-max & $+6.06^{*\dagger}$ & $+3.79^{*\dagger}$ & $+4.59^{*\dagger}$ & $+3.15^{*}$ \\
 & {\scriptsize$[+1.55,+11.69]$} & {\scriptsize$[+0.90,+7.17]$} & {\scriptsize$[+0.24,+9.75]$} & {\scriptsize$[+0.03,+6.50]$} \\
\bottomrule
\end{tabular*}
\begin{tablenotes}
\footnotesize
\item \textit{Notes:} This table reports the mean marginal decision influence $\mathrm{Use}(\ell)$ across the 12 sample firms, with 95\% confidence intervals from a firm-level bootstrap (10{,}000 draws) in brackets. The primary criterion is the pre-specified empirical null, which compares target effects with identical pseudo-insertions rather than with zero: $^{\dagger}$ marks cells significant under that criterion (permutation $p<0.05$; available for the primary and API arms, whose designs include pseudo-insertions; full $p$-values in Appendix Table~\ref{tab:app_permutation}). $^{*}$ marks the descriptive criterion that the bootstrap interval excludes zero; where the two disagree---most notably the primary model's 128k cell ($p=0.95$ against the empirical null)---the empirical-null reading governs. Six malformed API numeric responses are excluded; affected firm-cells require at least four of five neutral controls.
\end{tablenotes}
\end{threeparttable}
\end{table}

\clearpage
\begin{table}[p]
\centering
\begin{threeparttable}
\caption{Declining influence is not retrieval failure}
\label{tab:recognition}
\small
\begin{tabular*}{\textwidth}{@{\extracolsep{\fill}}lcccc}
\toprule
 & \multicolumn{4}{c}{Context length} \\
\cmidrule(lr){2-5}
 & 2k & 8k & 32k & 128k \\
\midrule
\multicolumn{5}{l}{\emph{Panel A: directed retrieval of the target facts (target present)}} \\
\addlinespace
Qwen3.5-9B-Base & 12/12 & 12/12 & 12/12 & 12/12 \\
gemma-4-12B-it & 12/12 & --- & --- & 10/10 \\
qwen3.8-max & 0.77 & 0.76 & 0.72 & 0.74 \\
\addlinespace
\multicolumn{5}{l}{\emph{Panel B: use as a share of its own 2k level (before-arrangement)}} \\
\addlinespace
Qwen3.5-9B-Base & 1.00 & 0.14 & 0.03 & 0.12 \\
gemma-4-12B-it & 1.00 & 0.37 & $-$0.25 & $-$0.06 \\
qwen3.8-max & 1.00 & 0.63 & 0.76 & 0.52 \\
\bottomrule
\end{tabular*}
\begin{tablenotes}
\footnotesize
\item \textit{Notes:} Panel A reports directed-retrieval performance on target filings: for open-weight models, the number of firms whose answers match the frozen fact keys over firms evaluated; for the API model, the mean per-item completeness score. The cross-family rows run retrieval at the design's two endpoint lengths. Neutral-control filings generate zero false-positive retrievals for Qwen models at every length; Llama and Gemma show 5--25\% false-positive rates, a family difference read through the target-minus-neutral retrieval contrast rather than raw pass rates. Panel B rescales each model's $\mathrm{Use}(\ell)$ by its own 2k value. Retrieval remains at or near ceiling at 128k while use declines toward---and in gemma-4 through---zero.
\end{tablenotes}
\end{threeparttable}
\end{table}

\clearpage
\begin{table}[p]
\centering
\begin{threeparttable}
\caption{Model capability moves the integration horizon outward}
\label{tab:scale}
\small
\begin{tabular*}{\textwidth}{@{\extracolsep{\fill}}lcccc}
\toprule
 & \multicolumn{4}{c}{$\mathrm{Use}(\ell)$, before-arrangement} \\
\cmidrule(lr){2-5}
Model & 2k & 8k & 32k & 128k \\
\midrule
Qwen3.5-2B & $+0.023^{*}$ & $+0.003$ & $+0.030^{*}$ & $+0.007$ \\
Qwen3.5-9B (post-trained) & $-0.018$ & $-0.051^{*}$ & $-0.047^{*}$ & $-0.030^{*}$ \\
Qwen3.5-9B-Base & $+0.032^{*}$ & $+0.005$ & $+0.001$ & $+0.004^{*}$ \\
Qwen3.5-27B & $+0.105^{*}$ & $+0.026^{*}$ & $+0.025^{*}$ & $+0.034^{*}$ \\
\bottomrule
\end{tabular*}
\begin{tablenotes}
\footnotesize
\item \textit{Notes:} $^{*}$ marks bootstrap intervals excluding zero, as in Table~\ref{tab:main_use}; the pre-specified empirical-null criterion is available for the 9B-Base row (Appendix Table~\ref{tab:app_permutation}). The capability claim rests on the endpoints: the 27B model is the only open-weight model whose use remains significant at every length, and its 128k use approximately equals the 9B models' 2k use. The 9B-Base row is the mechanism workhorse (its choice is dictated by the availability of pretrained sparse autoencoders, Section~\ref{sec:design}). The post-trained 9B exhibits an arrangement-dependent sign reversal discussed in the text (negative use in the before-arrangement shown here; positive in the after-arrangement; Appendix Table~\ref{tab:app_inversions}). The 2B model orders severity but discriminates within a compressed band around a strong sell prior (Appendix Table~\ref{tab:app_sevgate}); its non-monotone pattern reflects that compression.
\end{tablenotes}
\end{threeparttable}
\end{table}

\clearpage
\begin{table}[p]
\centering
\begin{threeparttable}
\caption{Exploratory extension: surgical redaction of twenty actual 10-K disclosures (capability-conditional replication)}
\label{tab:real_disclosure}
\small
\begin{tabular*}{\textwidth}{@{\extracolsep{\fill}}lcccc}
\toprule
Model & Use, excerpt & Use, full filing & Placebo edit & Retrieval, full \\
\midrule
27B (capable open-weight) & \begin{tabular}{@{}c@{}}$+0.037$\\{\scriptsize$[-0.003,+0.079]$}\end{tabular} & \begin{tabular}{@{}c@{}}$+0.000$\\{\scriptsize$[-0.003,+0.003]$}\end{tabular} & \begin{tabular}{@{}c@{}}$+0.001$\\{\scriptsize$[-0.001,+0.003]$}\end{tabular} & 20/20 \\
\quad{\footnotesize adverse-only subset (n=16)} & \begin{tabular}{@{}c@{}}$+0.042^{*}$\\{\scriptsize$[+0.003,+0.089]$}\end{tabular} &  &  &  \\
\addlinespace
9B-Base (workhorse) & \begin{tabular}{@{}c@{}}$-0.006$\\{\scriptsize$[-0.019,+0.008]$}\end{tabular} & \begin{tabular}{@{}c@{}}$-0.001$\\{\scriptsize$[-0.004,+0.003]$}\end{tabular} & \begin{tabular}{@{}c@{}}$+0.001$\\{\scriptsize$[-0.002,+0.004]$}\end{tabular} & 13/19 \\
\quad{\footnotesize adverse-only subset (n=16)} & \begin{tabular}{@{}c@{}}$-0.006$\\{\scriptsize$[-0.021,+0.010]$}\end{tabular} &  &  &  \\
\addlinespace
\bottomrule
\end{tabular*}
\begin{tablenotes}
\footnotesize
\item \textit{Notes:} Each of twenty registrants' actual filings (all filed after the models' training cutoffs) is measured verbatim and with its focal disclosure surgically replaced by a length-matched neutral paragraph; every other passage that repeats the disclosure's distinctive amounts is excised from both versions, so the contrast isolates the disclosure itself. Use is the sell-probability gap between the intact and redacted versions; the placebo column applies the same edit to an unrelated, token-matched passage. Excerpts are $\sim$2,000-token windows around the disclosure. Brackets are seeded firm-bootstrap 95\% intervals (10{,}000 draws); $^{*}$ marks intervals excluding zero. Retrieval is directed recognition scored against frozen answer keys on the complete filing.
\end{tablenotes}
\end{threeparttable}
\end{table}

\clearpage
\begin{table}[p]
\centering
\begin{threeparttable}
\caption{Transplanting one memory channel at a time: the running summary carries a large share of the disclosure's influence}
\label{tab:transplant}
\footnotesize
\setlength{\tabcolsep}{3pt}
\begin{tabular*}{\textwidth}{@{\extracolsep{\fill}}lccc}
\toprule
Quantity (decision-score units, 2{,}000 tokens) & Mean & 95\% CI & Firms $>0$ \\
\midrule
Influence of the disclosure (baseline) & $+0.191^{*}$ & {\scriptsize$[+0.110,+0.268]$} & 10/12 \\
\quad removed when only the running summary forgets it & $+0.124^{*}$ & {\scriptsize$[+0.064,+0.187]$} & 10/12 \\
\quad recreated when only the running summary retains it & $+0.085^{*}$ & {\scriptsize$[+0.026,+0.147]$} & 10/12 \\
\quad neutral-to-neutral sham transplant (magnitude bound) & $0.046$ & {\scriptsize (mean $|\Delta|$, n=24)} &  \\
\quad sham, signed per-firm mean (surgery bias) & $+0.008$ & {\scriptsize$[-0.010,+0.024]$} & 8/12 \\
\quad removal $-$ sham, firm-paired & $+0.116^{*}$ & {\scriptsize$[+0.060,+0.179]$} & 11/12 \\
\bottomrule
\end{tabular*}
\begin{tablenotes}
\footnotesize
\item \textit{Notes:} Target and neutral documents are byte-identical outside the disclosure's slot, so both can be processed to the end of the slot and their two forms of memory recombined: the lookup store (attention key--value cache) from one run with the running summary (the recurrent state of the model's linear-attention layers) from the other. Row two processes the target document while only the running summary comes from the neutral run; row three processes the neutral document while only the running summary comes from the target run. As a ratio of full-sample means, the transplant removes 65\% of the baseline contrast and recreates 44\%. Three identification caveats apply: the contrast uses a single neutral version rather than the design's mean of five; slot boundaries differ by a few tokens across versions, and a neutral-to-neutral sham transplant bounds that generic perturbation (fourth to sixth rows: swapping the channel between two neutral documents perturbs the readout, but its signed per-firm mean is indistinguishable from zero---the surgery is noisy but unbiased---and the firm-paired contrast of the real removal against the sham remains large and consistently positive, exact sign-flip $p=0.002$); and these ratios are therefore evidence that the running summary carries a large share of the disclosure's influence, not an additive channel decomposition. Brackets: seeded firm bootstrap, 95\%; $^{*}$ excludes zero.
\end{tablenotes}
\end{threeparttable}
\end{table}

\clearpage
\begin{table}[p]
\centering
\begin{threeparttable}
\caption{Reorganizing the decision process restores disclosure influence; additional computation does not}
\label{tab:remedies}
\small
\begin{tabular*}{\textwidth}{@{\extracolsep{\fill}}lcccc}
\toprule
Protocol & Use at 2k & Use at 128k & Retention (128k/2k) & Firms $>0$ at 128k \\
\midrule
Chunk-then-aggregate & $-0.001$ & $-0.001$ & --- & 5/12 \\
Baseline & $+0.032$ & $+0.004$ & 0.12 & 8/12 \\
Re-present excerpt & $+0.002$ & $+0.013$ & --- & 7/12 \\
Extract-then-decide & $+0.126$ & $+0.085$ & 0.67 & 12/12 \\
\bottomrule
\end{tabular*}
\begin{tablenotes}
\footnotesize
\item \textit{Notes:} Mean $\mathrm{Use}$ across 12 firms on Qwen3.5-9B-Base. The retention column is the within-protocol ratio of 128k use to 2k use---the share of the disclosure's short-context influence that survives at full length under each protocol; it is undefined for protocols whose 2k use is at the noise floor. Extract-then-decide raises retention from 12\% to 67\% (before-arrangement; both arrangements in Appendix Table~\ref{tab:app_remedies_full}). Extract-then-decide asks the model to answer the directed-retrieval question first and prepends its own answer to the decision prompt, keeping the filing in context. Chunk-then-aggregate replaces the filing with the model's own section-by-section notes, eliminating the disclosure's influence even at 2k, where the document fits in a single chunk. Re-present prepends a verbatim copy of the target paragraph. Extended-reasoning tiers on the production API model leave use unchanged at every effort level (Appendix Table~\ref{tab:app_thinking}).
\end{tablenotes}
\end{threeparttable}
\end{table}

\clearpage
\begin{table}[p]
\centering
\begin{threeparttable}
\caption{What makes extract-then-decide work: model authorship is unnecessary; targeted structured restatements perform best}
\label{tab:decomposition}
\footnotesize
\setlength{\tabcolsep}{3pt}
\begin{tabular*}{\textwidth}{@{\extracolsep{\fill}}lcccc}
\toprule
Injected before the decision & \multicolumn{2}{c}{Use at 2{,}000} & \multicolumn{2}{c}{Use at 128{,}000} \\
 & mean & firms $>0$ & mean & firms $>0$ \\
\midrule
Model's own targeted extraction (extract-then-decide) & $+0.113^{*}$ & 12/12 & $+0.055^{*}$ & 11/12 \\
Experimenter-written targeted extraction (answer key) & $+0.151^{*}$ & 11/12 & $+0.056^{*}$ & 12/12 \\
Another model's targeted extraction & $+0.050^{*}$ & 9/12 & $+0.013$ & 8/12 \\
Model's own verbatim copy of the passage & $+0.016$ & 8/12 & $+0.018^{*}$ & 10/12 \\
Verbatim re-presentation of the passage & $-0.002$ & 6/12 & $+0.004$ & 6/12 \\
\bottomrule
\end{tabular*}
\begin{tablenotes}
\footnotesize
\item \textit{Notes:} Every condition injects text into the same position with the same framing immediately before the decision question, with the full filing still in context (after-arrangement; both arrangements and confidence intervals in Appendix Table~\ref{tab:app_matched_full}). The first two rows are statistically indistinguishable: an experimenter-written extraction in the same structured form works as well as the model's own, so self-generation is not the operative ingredient. Another model's extraction (row three) recovers about a quarter of that effect at full length and its interval includes zero; the model's own words in unstructured verbatim form (row four) retain a small significant effect and re-presented source text (row five) none. The conditions jointly vary form, fidelity, and wording, so they identify that model authorship is unnecessary conditional on the structured targeted form---not which single ingredient mediates. Neutral cells run the identical pipeline, so each contrast nets out the injection itself.
\end{tablenotes}
\end{threeparttable}
\end{table}

\clearpage
\begin{table}[p]
\centering
\begin{threeparttable}
\caption{On complete real 10-K filings, workflow design decides whether the disclosure reaches the recommendation}
\label{tab:eco_remedies}
\small
\begin{tabular*}{\textwidth}{@{\extracolsep{\fill}}lccc}
\toprule
Protocol & Mean Use & 95\% CI & Firms $>0$ \\
\midrule
Direct full-filing reading & $-0.004^{*}$ & {\scriptsize$[-0.007,-0.000]$} & 5/12 \\
Chunk-then-aggregate (summary notes) & $-0.003$ & {\scriptsize$[-0.014,+0.008]$} & 6/12 \\
Extract-then-decide & $+0.043^{*}$ & {\scriptsize$[+0.026,+0.059]$} & 11/12 \\
\bottomrule
\end{tabular*}
\begin{tablenotes}
\footnotesize
\item \textit{Notes:} Twelve complete, surgically edited real filings (Section~\ref{sec:design}); Use is the sell-probability contrast between the disclosure-bearing version and the mean of five neutral versions. Direct reading of the full filing leaves the disclosure without influence, and chunked summarization cannot destroy what is already absent---its notes audit shows the disclosure's facts survive into the notes in 0 of 12 filings (7--15 chunks each). The direct-reading baseline is slightly negative rather than literally zero; extract-then-decide creates influence from that essentially nil baseline. Brackets: seeded firm bootstrap, 95\%; $^{*}$ excludes zero.
\end{tablenotes}
\end{threeparttable}
\end{table}

\clearpage
\clearpage

\clearpage
\clearpage

\clearpage

\clearpage

\clearpage

\clearpage
\appendix

\clearpage
\section{Materials and Stimulus Construction}\label{app:materials}

\subsection{Target Disclosures}
Each of the twelve sample firms receives one researcher-authored target
paragraph describing a firm-specific risk with verifiable quantities. We
restrict attention to fact classes that can be described completely within a
single passage---debt-covenant thresholds and measured headroom, litigation
settlement amounts, and indemnification caps---because such facts can be
added to a filing at one location without creating obligations to restate
other numbers. Every candidate paragraph passes a whole-filing coherence
audit before entering the sample: we check for conflicts with the three
primary financial statements, the related notes, management's discussion and
analysis, the risk-factor section, legal proceedings, and the audit opinion;
we check that the fact is consistent with the filing's timeline; and we ask
whether GAAP or SEC rules would require the fact, if true, to alter other
disclosures. A single discard rule governs the audit: if a candidate fact
would require edits in more than one location to be coherent, the candidate
is dropped rather than patched. Multi-point edits would destroy the
single-treatment interpretation of the design.

\subsection{Controls, Severity Ladder, and Filler}
For each firm we author five neutral replacement paragraphs of equal length
that occupy the target's slot without conveying decision-relevant
information; the marginal decision influence measure differences the target
condition against the mean of the five neutral conditions, absorbing any
deterministic effect of occupying the slot. Each firm additionally has a
seven-step severity ladder ($S_1$--$S_7$) of increasingly adverse versions of
the target fact. The full ladder runs at the 2k and 128k endpoints, with
anchor severities at 8k and 32k; the ladder calibrates the instrument
(Appendix Table~\ref{tab:app_sevgate}) and supports the
range-compression analysis in the main text.

Context length is manipulated with filler pools of genre-matched accounting
and legal text that is economically unrelated to the focal firm and its
risk. Pools are pre-assigned across firms. Because ``unrelated'' is a design
property rather than a linguistic absolute, filler is certified empirically:
we verify that pure-filler baselines move the level of the decision readout
(they make the models modestly more optimistic) but that this level shift is
common to target and neutral conditions and is therefore absorbed by
differencing. Documents are assembled at sentence boundaries under exact
token budgets, holding the focal mini-filing---the target or neutral
paragraph, its local headings, and adjacent text---byte-identical across
lengths. All materials are frozen in a registry with content hashes before
any measurement run.

\subsection{Ecological Arm}
The stylized documents purchase identification at the cost of realism, so we
validate the central retrieval--integration gap in complete, investor-visible 10-K
filings. For each firm the target paragraph is inserted into the real filing
at a recorded anchor location, and original passages that would contradict
the inserted fact are removed through machine-verified verbatim deletions
(the deletion registry holds 33 entries for the twelve filings, of which
30 are nonzero logical deletions covering 33 removed occurrences, taken
as the union across severity
levels so a single edited document serves all conditions). An assembly
audit verifies that every deletion matches exactly once, that no deleted
string survives, that the insertion appears exactly once, and that the edit
introduces no whitespace fingerprints (blank lines, trailing spaces, or
spacing patterns absent from the source); all filings pass all checks.
Results appear in Appendix Table~\ref{tab:app_ecological}.

\subsection{Real-Disclosure Redaction Sample}
\label{app:real_disclosures}
The redaction experiment of Table~\ref{tab:real_disclosure} reverses the
ecological arm's logic: rather than inserting an authored paragraph into a
real filing, it removes a real one. Candidate disclosures were screened
from EDGAR full-text search over 10-K filings filed February through
August 2026---after every studied model's training cutoff---in six
categories (covenant headroom or compliance status, debt maturity and
liquidity constraints, litigation settlements and exposures,
indemnification obligations, customer or supplier concentration, and
quantified regulatory liabilities). A candidate qualifies only if the
disclosure is self-contained in one passage, states at least two
verifiable quantities, and is not woven through the filing: a machine
prescreen requires the passage to match the normalized filing text exactly
once and counts every recurrence of its distinctive amounts elsewhere.
Twenty of twenty-five curated candidates pass; three fail because the
passage text itself recurs verbatim in the filing and two because the
curated transcription could not be aligned to the normalized text. For
each surviving firm we author a token-length-matched neutral replacement
(within ten percent of the passage, in the register of its section), a
directed-retrieval question, and frozen answer keys. Because a real
matter often echoes elsewhere in a filing, every passage outside the
focal disclosure that repeats its distinctive amounts is excised from
\emph{all} document versions (the excision registry holds 102 reviewed
sentence-level records, of which 100 are applicable and 99 are applied
as deletions in the current builds), so
the redaction contrast isolates the focal passage; without this
excision the redacted version can still answer retrieval questions
from the echoes, which is the failure the excision removes. The placebo
version replaces a sentence-aligned, token-matched prose block far from
the disclosure with the same neutral text. Excerpt versions take a
sentence-aligned window of roughly 2{,}000 tokens centered on the
disclosure slot. Retrieval on the redacted full filing doubles as a
memorization probe: a correct answer there can come only from residual
echoes or pretraining, and the three firms that trigger it are flagged
(their measured influence is thereby understated). Each disclosure is
labeled by predicted direction; the labels were assigned before the
27B extension was run. The four labeled
ambiguous include a going-concern doubt resolved by a post-year-end
financing and two settlements that resolve litigation overhangs, and the
sign heterogeneity the labels predict is visible in the firm-level
results.

\clearpage
\section{Measurement, Readouts, and Inference}\label{app:measurement}

\subsection{Decision Readouts}
For open-weight models the decision readout is deterministic. A fixed
decision prompt asks for exactly one of three action codes (buy, hold,
sell); we read the next-token logits of the three codes at the final
position and report the sell probability from a softmax over the three.
Because a bf16 output head quantizes logits at approximately 0.125---coarser
than several contrasts we measure---the three code logits are recomputed in
float32 from the final-layer hidden state. Commercial API models do not
expose logits, so the cross-arm readout is a unit-bearing numeric judgment
(the twelve-month default probability, in percentage points) elicited under
an identical document and prompt structure. Across the full two-model,
two-arrangement API archive, seven numeric responses are malformed and
unparseable---all from the production model, six of them in the
before-arrangement cells the main table reports; affected firm-cells are
retained when at least four of the five neutral controls parse.

\subsection{Retrieval Readout}
Directed retrieval is a separate call that asks what the filing says about
the target matter. Answers are scored against frozen per-firm fact keys with
alias groups, so paraphrases of an amount or status match without judgment
calls at scoring time. Two accommodations were required for the cross-family
rows: normalization of paraphrased section labels and stripping of
code-fence wrappers around otherwise exact structured answers. Both changes
were verified to leave the primary family's scores invariant (0 of more than
3{,}700 rescored results changed). Neutral-condition filings provide a
false-positive check; the primary family produces zero false-positive
retrievals at every length, and the cross-family rows produce 5--25\%,
a difference the differenced design absorbs.

\subsection{Inference}
The unit of inference is the firm-filing. Two procedures are used. First, a
pooled randomization test against an empirical null: for every cell we
create ten pseudo-target insertions that apply the identical insertion
machinery to irrelevant paragraphs; each of 2{,}000 permutation draws picks
one pseudo-insertion per firm, recomputes the pooled mean influence, and the
target statistic is compared with the centered draw distribution
(Appendix Table~\ref{tab:app_permutation}). This machinery is what licenses
interpreting long-context null results as loss of use rather than
instrument failure: the same device resolves clearly non-zero effects at
short lengths. Second, firm-level bootstrap confidence intervals (10{,}000
draws) for every model-length-arrangement cell reported in
Table~\ref{tab:main_use}.

\subsection{Computational Environment}
Because several measured contrasts are small, the runtime is pinned: a
single library version, a single kernel path for the hybrid-attention
layers, fixed numeric precision, and a fixed device policy for all paired
conditions. Hardware-induced numerical variation is treated as a property of
the environment to be held constant, not averaged over. During the study a
multi-GPU communication defect was identified on the compute platform; all
results produced under the affected configuration were discarded and
regenerated under a verified-correct configuration.

\clearpage
\section{Mechanism Methods}\label{app:mechanism_methods}

\subsection{Vocabulary-Space Probe}
The first mechanism instrument reads intermediate representations along the
model's own vocabulary directions using the probe of
\citet{gurnee2026verbalizable}. The probe is fit on a held-out calibration
set of prompts and then, for each experimental document, applied at two
positions: the final token of the target disclosure (the ``fact position'')
and the final token before the model's answer (the ``decision position'').
At the fact position we record the probe rank of a frozen set of risk words;
at the decision position, the rank of the three action codes. Retention
trajectories interpolate the readout at intermediate positions between the
two. Rank readouts are summarized by medians across firms.

Because the workhorse architecture interleaves fixed-size recurrent
(linear-attention) layers with full-attention layers---an architecture class
whose retention properties are themselves an object of study
\citep{jelassi2024repeat, arora2024zoology}---we validate the probe by
refitting on a twelve-layer set containing six adjacent linear/full pairs.
Readability tracks depth, not attention type: median probe ranks agree
within pairs (approximately 122{,}000 versus 75{,}000 at layers 7/8,
declining to 26 versus 116 at layers 27/28), so the main-text findings at
the decision-readout layer are not artifacts of layer type.

\subsection{Sparse-Autoencoder Instrument}
The second instrument uses the pretrained residual-stream sparse autoencoders
of \citet{qwen2026scope}, in the tradition of dictionary-learning
interpretability \citep{templeton2024scaling}. For each analyzed layer we
freeze a risk-feature set---the twenty-four features with the largest mean
activation contrast between risk and non-risk sentences in a labeled
concept corpus---before looking at any experimental readout. We then record
summed risk-feature activation at the fact and decision positions.
Reconstruction quality at the analyzed layers is adequate for
representation decomposition (explained variance 0.65--0.78; cosine
similarity 0.81--0.88) and is not treated as evidence of causal validity.
Two facts anchor the main text: at the encoding layer, fact-position risk
activation is strongly target-specific at every length---approximately
sixty times the neutral level at 2k and roughly thirty times at 32k and
128k---while its absolute level is flat in context length; and at the decision
position, target and neutral documents are indistinguishable in this basis
at every length, including lengths at which behavior responds. The
decision position therefore appears to carry a readable pointer to the
pending decision rather than the disclosure's conceptual content, an
arrangement consistent with binding- and pointer-based accounts of
in-context reference \citep{feng2024binding, variengien2023look} and with
readout-time retrieval via attention \citep{geva2023dissecting,
olsson2022induction}.

\clearpage
\section{Causal Interventions}\label{app:interventions}

We attempted to convert the descriptive mechanism picture into causal
claims with three intervention designs, each with matched controls
(Appendix Table~\ref{tab:app_interventions}).

First, decision-position subspace patching: a rank-8 orthonormal basis for
the risk subspace is formed from the singular vectors of the frozen SAE
decoder directions, and the risk component of the decision-position
residual is restored (long context) or deleted (short context), with
rank- and norm-matched random subspaces, an unrelated-concept subspace, and
a non-workspace layer as controls. Second, fact-position subspace ablation:
the same basis is projected out of the residual stream across the
disclosure's entire token span at the encoding layer, in the lineage of
position-targeted causal tracing \citep{meng2022locating} and
sparse-feature ablation \citep{marks2025sparse}. Third, a short-to-long
transplant: the decision-position difference between target and mean
neutral representations at 2k---a difference-in-means direction in the
tradition of activation steering \citep{turner2023steering,
arditi2024refusal} and cross-context request patching
\citep{variengien2023look}---is added at the 128k decision position at two
magnitudes, with a norm-matched random direction and, critically, the same
patch applied to a neutral document in which there is nothing to
dereference.

Instrument resolution was verified rather than assumed: at the scale of the
sought effects a bf16 readout quantizes distinct conditions to
bitwise-identical effects, so all interventions use a float32 readout. Under the
verified instrument all three designs produce nulls with working controls:
ablation effects do not exceed matched random ablations
($t=-1.38$ across firms), transplants produce no dose response and no
target/neutral asymmetry, and a wrong-layer control that reached $t=+2.41$
in the wrong direction calibrates the noise scale of the test battery.
These nulls establish boundary conditions---the decision is not causally
carried by decision-position conceptual content, by a single-layer
fact-position risk subspace, or by transplantation of the
decision-position direction---and, in retrospect, they are the expected
signature of a carrier that is distributed and compressed rather than
localized. We treat all subspace-intervention results with interpretive
caution: apparently positive effects can arise through causally
disconnected pathways \citep{makelov2024subspace}, while nulls can
reflect downstream recomputation rather than the absence of mediation
\citep{vig2020investigating}. The two designs below address the
remaining routes: the attention battery bounds the lookup channel, and
the memory-channel transplant provides the study's strongest causal
evidence, implicating the recurrent running summary as a major carrier
without claiming an exclusive one.

\subsection{Attention-Access Interventions}
\label{app:attn_methods}
Decision-time attention is measured without materializing any full
attention matrix: the document is prefilled in cache-preserving blocks
exactly as the frozen decision readout runs, and the final token is then
forwarded alone against the cache, so each attention layer returns the
decision position's full row of weights. Localization contrasts the mass
on the disclosure span in the target document with the mass on the same
slot's neutral replacement in the neutral document (a content-selectivity
measure), with a length-matched control span between the disclosure and
the query. Access interventions zero and renormalize the post-softmax
weights from the listed query positions to the listed key span for the
listed heads, following the restore/destroy logic of causal tracing
\citep{meng2022locating} and head-level circuit analysis
\citep{wang2023interpretability}, with random-head, wrong-span, and
scope-ladder controls chosen per the patching-practice recommendations of
\citet{zhang2024towards}. The original battery's all-heads conditions
lacked their matched shams; the completing battery adds them. The
operative matched control is the identical all-heads blackout applied to
the neutral document's slot span, which is cleanly null, and the
identifying quantity is the difference-in-differences between the target
and neutral slot conditions. A second sham on a content-free control
span proved unstable across firms (individual-firm effects range over
several decision-score units) and is reported as a failed control, not
used in the identification. Every condition, baselines
included, runs under
one attention implementation with an explicit causal mask, so conditions
differ from their baseline only by the intervention; all decision logits
are recomputed in float32.

\subsection{Memory-Channel Transplant}
\label{app:transplant_methods}
The hybrid architecture keeps two caches during prefill: per-layer
key--value stores for its full-attention layers and per-layer recurrent
states (a convolutional buffer and a gated-delta-rule state
\citep{yang2025gated}) for its linear-attention layers. Because target
and neutral documents are byte-identical outside the disclosure slot,
both are prefilled to their slot-end boundary and the two channels are
recombined by copying the linear layers' states from one run into the
other run's cache; the hybrid then processes the remainder of the
document---itself byte-identical across versions---and the decision is
read out in float32. The transplanted state carries no positional index,
and the key--value store defines the token stream, so each hybrid
continues with the suffix of the document whose lookup store it keeps.
Three design limitations qualify the estimates. The contrast is taken
against a single neutral version rather than the main design's mean of
five. The target and neutral slots differ slightly in tokenized length
($-9$ to $+11$ tokens across firms; mean absolute difference 5.25), so
the splice point is not perfectly aligned; a neutral-to-neutral sham
transplant---recombining the channels of two neutral versions, where no
content difference exists---bounds the generic boundary and
cache-compatibility perturbation this could introduce. And the two
channel effects need not be additive, so the removal and recreation
ratios are not a decomposition. One firm (of twelve) responds to neither
transplant direction despite a positive baseline contrast; the routing
of its influence is not identified, and it is reported unmodified in all
aggregates. In the
pure-attention comparison architecture there is no recurrent channel to
transplant; there, the all-heads access intervention is the operative
test, and its direction is consistent with attention carrying the load
(seven of twelve firms negative) but its magnitude is not sharply
estimated because that family's short-context baseline effect is itself
small---we therefore claim the cross-architecture replication for the
localization signature, not for a carrier share. The signature itself
also differs in geography: the workhorse's most selective heads
concentrate in late layers, while the comparison architecture's top five
sit at layers 0, 0, 30, 0, and 28---content-selective reading exists in
both, but where it lives is architecture-specific.

\clearpage
\section{Robustness and Additional Results}\label{app:robustness}

\subsection{Capability Gate}
A model that cannot order disclosure severities in a short context cannot
carry length-decay interpretation. Appendix Table~\ref{tab:app_sevgate}
reports the gate: the post-trained 9B and the base 9B pass with wide
severity ranges; the 2B passes marginally, discriminating within a
compressed band around a strong sell prior; and the full-attention control
model fails outright (a two-percentage-point range, rank correlation near
zero) and is excluded from decay interpretation. Post-training
approximately doubles severity discrimination on identical stimuli.

\subsection{Arrangement Contrast}
Placing filler before versus after the focal filing changes the distance
between disclosure and decision. Appendix Table~\ref{tab:app_h3} shows the
behavioral arrangement contrast is a calibrated null: at every length the
before-minus-after gap is statistically indistinguishable from the same
gap computed on pseudo-target insertions. The retention interpretation in
the main text rests on the internal-representation evidence and the
workflow gradient, not on this contrast.

\subsection{Sign Inversions}
Appendix Table~\ref{tab:app_inversions} collects cells in which the risk
disclosure makes models marginally \emph{less} likely to sell: one
mid-length cell in one family, one long-context cell in another, and an
arrangement-dependent flip in the post-trained 9B. Roughly two
false-positive stars are expected by chance across the forty cells tested,
so we report these as suggestive; the cross-family echo at adjacent lengths
is why they are reported at all.

\subsection{Repetition and Semantic Interference}
Controlled repetition of the disclosure ($k=2,4$; total length and final
position fixed) helps partially and unevenly: fourfold repetition exceeds
the single-occurrence level in the pooled length cells, twofold repetition
does not improve every cell, and neither restores
long-context use---fourfold repetition at 128k recovers less than one third
of a single mention's 2k influence. Same-risk filler---disclosures of
similar risks from other firms---separates from unrelated filler only on
the production API model, where it reduces use at 2k and turns it negative
at 8k; the local workhorse shows no separation. Semantic interference is
therefore real but capability-dependent, and it is not the primary driver
of the main effect, which obtains with economically unrelated filler.

\subsection{Out-of-Sample API Arm and Workflow Details}
A dated API snapshot that predates the sample filings replicates the
retrieval result (flat directed-retrieval performance, zero neutral
false positives) but its numeric readout is too noisy for firm-level
behavioral claims; we retain it solely as an out-of-sample retrieval
witness. Appendix Table~\ref{tab:app_remedies_full} reports the full
workflow matrix in both arrangements; the extract-then-decide workflow's
internal ratio (128k to 2k of roughly 0.5--0.7) shows residual length
erosion inside the workflow itself. Appendix Table~\ref{tab:app_thinking}
shows extended-reasoning tiers do not restore use at any effort level and
reduce measured use at 2k, consistent with length effects that persist
despite successful retrieval \citep{du2025context} and with
benchmark evidence that long-context capacity generalizes poorly
\citep{liu2024lost, hsieh2024ruler, yen2025helmet, levy2024same};
positional-attention accounts and calibration remedies
\citep{hsieh2024found} suggest one route by which such effects arise.
Finally, our bounded, generic implementation of a chunked, note-based
pipeline eliminates the
disclosure's influence entirely in our setting, even at 2k where the
document fits in one chunk---the practical warning the workflow-architecture
section develops.

\clearpage

\begin{table}[p]
\centering
\begin{threeparttable}
\caption{Severity capability gate at 2k tokens}
\label{tab:app_sevgate}
\small
\begin{tabular*}{\textwidth}{@{\extracolsep{\fill}}lcccc}
\toprule
Model & Arrangement & $S_7-S_1$ range & Spearman $\rho$ & Verdict \\
\midrule
Qwen3.5-9B (post-trained) & before & $+0.42$ & $+0.48$ & pass \\
Qwen3.5-9B (post-trained) & after & $+0.34$ & $+0.38$ & pass \\
Qwen3.5-9B-Base & before & $+0.23$ & $+0.54$ & pass \\
Qwen3.5-2B & before & $+0.09$ & $+0.43$ & marginal \\
Qwen3-8B-Base (full attn.) & both & $+0.02$ / $-0.00$ & $+0.10$ / $-0.09$ & fail \\
\bottomrule
\end{tabular*}
\begin{tablenotes}
\footnotesize
\item \textit{Notes:} Mean sell probability by severity level (S1--S7) at 2k tokens, 12 firms pooled. A model that cannot order severities at 2k cannot carry length-decay interpretation; the full-attention control fails this gate and is excluded from the main analysis. Post-training approximately doubles severity discrimination relative to the base model on identical stimuli.
\end{tablenotes}
\end{threeparttable}
\end{table}

\begin{table}[p]
\centering
\begin{threeparttable}
\caption{Pooled randomization inference against the empirical insertion null}
\label{tab:app_permutation}
\small
\begin{tabular*}{\textwidth}{@{\extracolsep{\fill}}llcccc}
\toprule
 & & \multicolumn{4}{c}{Two-sided permutation $p$ (2{,}000 draws)} \\
\cmidrule(lr){3-6}
Model & Arrangement & 2k & 8k & 32k & 128k \\
\midrule
Qwen3.5-9B-Base & before & 0.025 & 0.8481 & 0.5707 & 0.947 \\
Qwen3.5-9B-Base & after & 0.0015 & 0.0385 & 0.935 & 0.8026 \\
qwen3.8-max & before & 0.0005 & 0.0005 & 0.0005 & 0.1079 \\
qwen3.8-max & after & 0.0005 & 0.0005 & 0.0965 & 0.4923 \\
\bottomrule
\end{tabular*}
\begin{tablenotes}
\footnotesize
\item \textit{Notes:} The null distribution draws one pseudo-target insertion ($P_1..P_{10}$) per firm and recomputes the pooled mean use; the target statistic is compared with 2{,}000 centered draws. The pseudo-insertion null has a small positive mean, which the centering removes; this machinery is what licenses interpreting the 128k null as loss of use rather than instrument failure. Numeric-task pseudo insertions parse at 60--100\% per firm; a firm contributes when at least five of ten pseudo draws parse.
\end{tablenotes}
\end{threeparttable}
\end{table}

\begin{table}[p]
\centering
\begin{threeparttable}
\caption{Full remedy results, both arrangements}
\label{tab:app_remedies_full}
\small
\begin{tabular*}{\textwidth}{@{\extracolsep{\fill}}lcccc}
\toprule
Protocol & 2k before & 2k after & 128k before & 128k after \\
\midrule
Baseline & $+0.032$ & $+0.025$ & $+0.004$ & $+0.001$ \\
R1 chunk-then-aggregate & $-0.001$ & $-0.002$ & $-0.001$ & $-0.002$ \\
R2 extract-then-decide & $+0.126$ & $+0.113$ & $+0.085$ & $+0.055$ \\
R3 re-present excerpt & $+0.002$ & $-0.002$ & $+0.013$ & $+0.004$ \\
\bottomrule
\end{tabular*}
\begin{tablenotes}
\footnotesize
\item \textit{Notes:} Mean $\mathrm{Use}$ across 12 firms, Qwen3.5-9B-Base. R2's within-protocol ratio (128k/2k $\approx$ 0.5--0.7) shows residual length erosion inside the remedy. R1 note text is archived verbatim, and the notes audit separating extraction failure from decision-stage neglect appears in Appendix Table~\ref{tab:app_r1_audit}.
\end{tablenotes}
\end{threeparttable}
\end{table}

\begin{table}[p]
\centering
\begin{threeparttable}
\caption{Extended reasoning does not restore use (production API model)}
\label{tab:app_thinking}
\small
\begin{tabular*}{\textwidth}{@{\extracolsep{\fill}}lcc}
\toprule
Configuration & Use at 2k (pp) & Use at 128k (pp) \\
\midrule
No reasoning (primary) & $+6.1$ & $+3.1$ \\
Reasoning, low effort & $+1.9$ & $+2.2$ \\
Reasoning, extra-high effort & $+2.0$ & $+0.2$ \\
\bottomrule
\end{tabular*}
\begin{tablenotes}
\footnotesize
\item \textit{Notes:} qwen3.8-max numeric-judgment readout, before-arrangement, mean across firms. Enabling extended reasoning reduces measured use at 2k and does not restore it at 128k at any effort tier.
\end{tablenotes}
\end{threeparttable}
\end{table}

\begin{table}[p]
\centering
\begin{threeparttable}
\caption{The arrangement contrast is a calibrated null}
\label{tab:app_h3}
\small
\begin{tabular*}{\textwidth}{@{\extracolsep{\fill}}lccccc}
\toprule
Length & Use before & Use after & Gap & Pseudo gap & $z$ vs.\ pseudo \\
\midrule
2k & $+0.032$ & $+0.025$ & $+0.007$ & $+0.008$ & $-0.09$ \\
8k & $+0.005$ & $+0.006$ & $-0.002$ & $+0.004$ & $-0.59$ \\
32k & $+0.001$ & $+0.001$ & $-0.001$ & $-0.003$ & $+0.45$ \\
128k & $+0.004$ & $+0.001$ & $+0.003$ & $+0.003$ & $-0.01$ \\
\bottomrule
\end{tabular*}
\begin{tablenotes}
\footnotesize
\item \textit{Notes:} The before$-$after gap in use is compared with the identical paired contrast computed on pseudo-target insertions. At every length the behavioral arrangement contrast is statistically indistinguishable from the insertion-machinery artifact; the retention mechanism is carried by the internal-representation evidence, not by this contrast.
\end{tablenotes}
\end{threeparttable}
\end{table}

\begin{table}[p]
\centering
\begin{threeparttable}
\caption{Sign inversions at long context}
\label{tab:app_inversions}
\small
\begin{tabular*}{\textwidth}{@{\extracolsep{\fill}}lccc}
\toprule
Model & Cell & Mean use & 95\% CI \\
\midrule
gemma-4-12B-it & 32k, before & $-0.012$ & $[-0.018, -0.006]$ \\
Llama-3.1-8B & 128k, before & $-0.006$ & $[-0.011, -0.001]$ \\
Qwen3.5-9B (post) & 8k--128k, before & $-0.051$ to $-0.030$ & all exclude 0 \\
Qwen3.5-9B (post) & 8k--32k, after & $+0.009$ to $+0.019$ & all exclude 0 \\
\bottomrule
\end{tabular*}
\begin{tablenotes}
\footnotesize
\item \textit{Notes:} Negative entries indicate that the risk disclosure makes the model marginally \emph{less} likely to sell. Approximately two false-positive stars are expected by chance across the forty cells tested; the cross-family echo at adjacent lengths, and the arrangement-dependent flip in the post-trained 9B, motivate reporting these as suggestive rather than confirmed.
\end{tablenotes}
\end{threeparttable}
\end{table}

\begin{table}[p]
\centering
\begin{threeparttable}
\caption{Causal-intervention boundary: a resolution-verified null}
\label{tab:app_interventions}
\small
\begin{tabular*}{\textwidth}{@{\extracolsep{\fill}}lll}
\toprule
Design & Statistic & Verdict \\
\midrule
Decision-position risk-subspace patch & effects $\le 0.5$pp, $=$ controls & null \\
Fact-position subspace ablation (L16) & risk$-$random $=-0.0024$, $t=-1.38$ & null \\
Wrong-layer control (L24) & $t=+2.41$, wrong direction & calibrates noise \\
Short$\to$long pointer transplant & $\alpha{=}1$: $+0.0002$; $\alpha{=}2$: $+0.0001$ & null \\
Random-direction / no-referent controls & $+0.0005$ / $+0.0005$ & --- \\
\bottomrule
\end{tabular*}
\begin{tablenotes}
\footnotesize
\item \textit{Notes:} All interventions use an fp32 readout, whose resolution is verified for the effects sought; a bf16 readout quantizes them at the logit step. Three mediation routes are excluded: decision-position conceptual content, single-layer fact-position subspace content (subject to recomputation by later layers), and transplantation of the decision-position direction. These nulls motivated the direct tests of the two memory channels---the attention-access battery and the memory-channel transplants---reported in the appendices that follow.
\end{tablenotes}
\end{threeparttable}
\end{table}

\begin{table}[p]
\centering
\begin{threeparttable}
\caption{Ecological validation: surgically edited full 10-K filings}
\label{tab:app_ecological}
\small
\begin{tabular*}{\textwidth}{@{\extracolsep{\fill}}lc}
\toprule
Quantity & Value \\
\midrule
Firms with complete results & 12 \\
Deletion-registry entries (union across severity levels) & 33 \\
\quad of which nonzero logical deletions & 30 \\
\quad occurrences removed across the twelve filings & 33 \\
Whitespace-fingerprint failures after assembly audit & 0 \\
Directed-retrieval completeness (target present) & 0.931 \\
Marginal decision influence & $-0.0037$ \\
\bottomrule
\end{tabular*}
\begin{tablenotes}
\footnotesize
\item \textit{Notes:} Real filings with the target risk surgically inserted (and conflicting original passages removed with recorded, machine-verified edits). The retrieval--use dissociation replicates in complete, investor-visible filings: the model retrieves the inserted disclosure almost perfectly and its investment judgment does not respond.
\end{tablenotes}
\end{threeparttable}
\end{table}

\begin{table}[p]
\centering
\begin{threeparttable}
\caption{Chunk-then-aggregate notes audit: the disclosure is evicted at extraction}
\label{tab:app_r1_audit}
\small
\begin{tabular*}{\textwidth}{@{\extracolsep{\fill}}lcccc}
\toprule
Length & Facts fully in notes & Partially & Absent & Mean use given absent \\
\midrule
2k & 0 & 0 & 24 & $-0.0011$ \\
128k & 1 & 2 & 21 & $-0.0015$ \\
\bottomrule
\end{tabular*}
\begin{tablenotes}
\footnotesize
\item \textit{Notes:} Firm-arrangement cells on target filings, notes archived verbatim and scored against the frozen retrieval fact keys (alias groups). Notes are generated under a per-segment token budget typical of practical pipelines; they routinely transcribe facts from the head of each segment until the budget binds (many truncate mid-sentence) and the single-point risk disclosure is evicted before the decision stage ever sees it. The audit attributes the protocol's zero influence to the extraction stage, not to decision-stage neglect of extracted facts.
\end{tablenotes}
\end{threeparttable}
\end{table}

\begin{table}[p]
\centering
\begin{threeparttable}
\caption{The pre-specified primary contrast, tested directly: within-firm decline in disclosure influence from 2{,}000 to 128{,}000 tokens}
\label{tab:app_paired_decline}
\footnotesize
\setlength{\tabcolsep}{3pt}
\begin{tabular*}{\textwidth}{@{\extracolsep{\fill}}lcccc}
\toprule
Model / arrangement & Mean decline & 95\% CI & Firms declining & Exact $p$ \\
\midrule
9B-Base, filler before & $+0.028$ & {\scriptsize$[+0.010,+0.046]$} & 8/12 & 0.0132 \\
9B-Base, filler after & $+0.024$ & {\scriptsize$[+0.014,+0.033]$} & 10/12 & 0.0024 \\
9B-Base, arrangements pooled & $+0.026$ & {\scriptsize$[+0.013,+0.038]$} & 10/12 & 0.0044 \\
Llama-3.1-8B, pooled & $+0.015$ & {\scriptsize$[+0.010,+0.022]$} & 11/12 & 0.0010 \\
Gemma-4-12B, pooled & $+0.040$ & {\scriptsize$[+0.007,+0.082]$} & 9/12 & 0.0352 \\
Production API (default prob., pp), pooled & $+2.47$ & {\scriptsize$[-0.65,+6.13]$} & 8/12 & 0.1992 \\
\bottomrule
\end{tabular*}
\begin{tablenotes}
\footnotesize
\item \textit{Notes:} Each firm contributes one paired difference $\mathrm{Use}_{2k}-\mathrm{Use}_{128k}$ (averaged within firm when arrangements are pooled). Open-weight rows are in sell-probability units; the production-API row is in percentage points of stated default probability. The exact $p$ enumerates all $2^{12}$ sign flips of the firm-level differences. The production-API decline is not statistically resolved and its curve should be read as a point-estimate pattern.
\end{tablenotes}
\end{threeparttable}
\end{table}

\begin{table}[p]
\centering
\begin{threeparttable}
\caption{The decline in disclosure influence replicates on a second economic judgment: probability of a credit-rating downgrade}
\label{tab:app_second_dv}
\small
\begin{tabular*}{\textwidth}{@{\extracolsep{\fill}}lcccc}
\toprule
Context length & \multicolumn{2}{c}{Use of the disclosure} & \multicolumn{2}{c}{Neutral dispersion} \\
 & filler before & filler after & before & after \\
\midrule
2,000 & \begin{tabular}{@{}c@{}}$+0.041^{*}$\\{\scriptsize$[+0.012,+0.074]$}\end{tabular} & \begin{tabular}{@{}c@{}}$+0.049^{*}$\\{\scriptsize$[+0.028,+0.075]$}\end{tabular} & 0.010 & 0.007 \\
8,000 & \begin{tabular}{@{}c@{}}$+0.026^{*}$\\{\scriptsize$[+0.008,+0.047]$}\end{tabular} & \begin{tabular}{@{}c@{}}$+0.014^{*}$\\{\scriptsize$[+0.006,+0.022]$}\end{tabular} & 0.013 & 0.005 \\
32,000 & \begin{tabular}{@{}c@{}}$+0.012$\\{\scriptsize$[-0.004,+0.029]$}\end{tabular} & \begin{tabular}{@{}c@{}}$+0.006$\\{\scriptsize$[-0.001,+0.013]$}\end{tabular} & 0.010 & 0.004 \\
128,000 & \begin{tabular}{@{}c@{}}$+0.003$\\{\scriptsize$[-0.010,+0.016]$}\end{tabular} & \begin{tabular}{@{}c@{}}$+0.006^{*}$\\{\scriptsize$[+0.002,+0.010]$}\end{tabular} & 0.008 & 0.004 \\
\bottomrule
\end{tabular*}
\begin{tablenotes}
\footnotesize
\item \textit{Notes:} One-token restricted-softmax readout of the probability that the model assesses a credit-rating downgrade as likely, on the workhorse model; Use is the target-minus-neutral contrast in this probability, and neutral dispersion is the standard deviation across the five neutral versions. Brackets: seeded firm bootstrap, 95\%, $^{*}$ excludes zero.
\end{tablenotes}
\end{threeparttable}
\end{table}

\begin{table}[p]
\centering
\begin{threeparttable}
\caption{Disclosure influence by total context and disclosure-to-decision distance}
\label{tab:app_factorial}
\small
\begin{tabular*}{\textwidth}{@{\extracolsep{\fill}}lcccc}
\toprule
Total context & \multicolumn{4}{c}{Disclosure-to-decision distance (tokens)} \\
 & 500--1{,}000 & 1{,}500--8{,}000 & 24{,}000 & 64{,}000 \\
\midrule
4,000 & \begin{tabular}{@{}c@{}}$+0.010^{*}$\\{\scriptsize$[+0.007,+0.014]$}\end{tabular} & \begin{tabular}{@{}c@{}}$+0.014^{*}$\\{\scriptsize$[+0.008,+0.019]$}\end{tabular} & --- & --- \\
8,000 & \begin{tabular}{@{}c@{}}$+0.007^{*}$\\{\scriptsize$[+0.000,+0.014]$}\end{tabular} & \begin{tabular}{@{}c@{}}$+0.007^{*}$\\{\scriptsize$[+0.003,+0.010]$}\end{tabular} & --- & --- \\
32,000 & \begin{tabular}{@{}c@{}}$-0.002$\\{\scriptsize$[-0.005,+0.002]$}\end{tabular} & \begin{tabular}{@{}c@{}}$+0.001$\\{\scriptsize$[-0.002,+0.004]$}\end{tabular} & \begin{tabular}{@{}c@{}}$+0.001$\\{\scriptsize$[-0.004,+0.005]$}\end{tabular} & --- \\
128,000 & \begin{tabular}{@{}c@{}}$-0.001$\\{\scriptsize$[-0.003,+0.001]$}\end{tabular} & \begin{tabular}{@{}c@{}}$+0.001$\\{\scriptsize$[-0.001,+0.003]$}\end{tabular} & \begin{tabular}{@{}c@{}}$+0.001$\\{\scriptsize$[-0.000,+0.003]$}\end{tabular} & \begin{tabular}{@{}c@{}}$+0.001$\\{\scriptsize$[-0.001,+0.003]$}\end{tabular} \\
\bottomrule
\end{tabular*}
\begin{tablenotes}
\footnotesize
\item \textit{Notes:} The same filler is split around the fixed focal-firm filing so that the disclosure sits at the stated distance from the decision question while total context is held fixed; the filler union is byte-identical across cells of a row; column headers give the distance range realized at each total. The 4{,}000- and 8{,}000-token rows sit below the failure horizon; there, moving the disclosure from 1{,}000 to 5{,}500 tokens before the decision at fixed 8{,}000-token total changes use by less than $0.0001$ in absolute value (exact sign-flip $p=0.99$), while raising total context from 8{,}000 to 32{,}000 at fixed 1{,}000-token distance lowers use by $0.008$ ($p=0.03$)---total load matters in the tested cells, a moderate distance change does not. At 32{,}000 tokens and beyond, all cells sit at the noise floor: moving the disclosure next to the decision does not rescue it. Brackets: seeded firm bootstrap, 95\%.
\end{tablenotes}
\end{threeparttable}
\end{table}

\begin{table}[p]
\centering
\begin{threeparttable}
\caption{A low-cost frontier-generation commercial system, held at its minimal reasoning setting}
\label{tab:app_frontier}
\footnotesize
\setlength{\tabcolsep}{3pt}
\begin{tabular*}{\textwidth}{@{\extracolsep{\fill}}lccccc}
\toprule
Model / length & Use (default, pp) & Use (P(sell)) & Retrieval & Sev.\ span (pp) & Neutral disp.\ (pp) \\
\midrule
\multicolumn{6}{l}{\emph{Gemini 3.1 Flash-Lite}} \\
\quad 2,000 & \begin{tabular}{@{}c@{}}$+5.74^{*}$\\{\scriptsize$[+1.52,+10.71]$}\end{tabular} & $+0.165^{*}$ & 12/12 & 45.0 & 3.71 \\
\quad 32,000 & \begin{tabular}{@{}c@{}}$+5.28^{*}$\\{\scriptsize$[+0.27,+10.88]$}\end{tabular} & $+0.042$ & 7/12 &  & 7.39 \\
\quad 128,000 & \begin{tabular}{@{}c@{}}$-7.17$\\{\scriptsize$[-18.53,+1.26]$}\end{tabular} & $+0.029$ & 7/12 & 18.0 & 8.29 \\
\addlinespace
\bottomrule
\end{tabular*}
\begin{tablenotes}
\footnotesize
\item \textit{Notes:} Behavior-only commercial arm on the frozen stimuli (after-arrangement), batch inference, reasoning fixed at the lowest setting the provider allows (low; the API does not expose a fully disabled mode). Use (default probability) is the target-minus-neutral contrast in the stated 12-month default probability, in percentage points; Use (P(sell)) is the same contrast in the share of sampled recommendations that are SELL (10 samples on the Qwen arm, 4 on Gemini---the provider cap). Severity span is the range of mean stated default probability across the seven-step severity ladder. Brackets: seeded firm bootstrap, 95\%; $^{*}$ excludes zero.
\end{tablenotes}
\end{threeparttable}
\end{table}

\begin{table}[p]
\centering
\begin{threeparttable}
\caption{Decision-time reading heads: content-selective attention to the disclosure exists at short length and collapses at full length, in both architectures}
\label{tab:app_attn_locate}
\footnotesize
\setlength{\tabcolsep}{3pt}
\begin{tabular*}{\textwidth}{@{\extracolsep{\fill}}lccccc}
\toprule
Layer.head & Selectivity, 2k & Consistency & Mass on target, 2k & Mass on target, 128k & Selectivity, 128k \\
\midrule
\multicolumn{6}{l}{\emph{Qwen3.5-9B-Base (hybrid)}} \\
\quad L27.H4 & $+0.0112$ & 11/12 & 0.0149 & 0.0006 & $+0.0005$ \\
\quad L3.H14 & $+0.0105$ & 12/12 & 0.0370 & 0.0006 & $+0.0001$ \\
\quad L27.H9 & $+0.0074$ & 12/12 & 0.0148 & 0.0002 & $+0.0001$ \\
\quad L27.H6 & $+0.0071$ & 8/12 & 0.0159 & 0.0004 & $+0.0002$ \\
\quad L27.H3 & $+0.0066$ & 11/12 & 0.0117 & 0.0002 & $+0.0001$ \\
\addlinespace
\multicolumn{6}{l}{\emph{Llama-3.1-8B (full attention)}} \\
\quad L0.H29 & $+0.0294$ & 11/12 & 0.0541 & 0.0003 & $+0.0000$ \\
\quad L0.H28 & $+0.0166$ & 9/12 & 0.0659 & 0.0010 & $+0.0000$ \\
\quad L30.H29 & $+0.0113$ & 12/12 & 0.0177 & 0.0004 & $+0.0002$ \\
\quad L0.H30 & $+0.0104$ & 8/12 & 0.0318 & 0.0006 & $+0.0000$ \\
\quad L28.H24 & $+0.0100$ & 11/12 & 0.0167 & 0.0007 & $+0.0005$ \\
\addlinespace
\bottomrule
\end{tabular*}
\begin{tablenotes}
\footnotesize
\item \textit{Notes:} Attention mass is measured from the final (decision) position over the whole document in a single-token forward pass against the prefilled cache. Selectivity is the mass a head places on the disclosure span in the target document minus the mass it places on the same slot's neutral replacement in the neutral document---the content-driven component of reading. Consistency counts firms with positive selectivity. The five most selective heads per architecture are shown.
\end{tablenotes}
\end{threeparttable}
\end{table}

\begin{table}[p]
\centering
\begin{threeparttable}
\caption{Severing attention access to the disclosure: the scope ladder}
\label{tab:app_attn_intervene}
\footnotesize
\setlength{\tabcolsep}{3pt}
\begin{tabular*}{\textwidth}{@{\extracolsep{\fill}}lccc}
\toprule
Condition (2{,}000 tokens) & Mean $\Delta$ & 95\% CI & Firms $<0$ \\
\midrule
Selected heads, disclosure span, query window & $+0.040$ & {\scriptsize$[-0.022,+0.134]$} & 4/12 \\
Random heads, disclosure span, query window & $+0.029$ & {\scriptsize$[-0.033,+0.129]$} & 5/12 \\
Selected heads, control span, query window & $+0.049$ & {\scriptsize$[-0.049,+0.175]$} & 8/12 \\
All heads, disclosure span, query window & $-0.054$ & {\scriptsize$[-0.140,+0.054]$} & 11/12 \\
All heads, disclosure span, every position after it & $-0.084^{*}$ & {\scriptsize$[-0.151,-0.013]$} & 11/12 \\
\midrule
\multicolumn{4}{l}{\emph{Query-window donor patches at 128{,}000 tokens ($\Delta$ vs.\ its base)}} \\
Restore: donor from target run & $+0.206^{*}$ & {\scriptsize$[+0.045,+0.353]$} & 10/12 $>0$ \\
Restore: donor from neutral run & $+0.184^{*}$ & {\scriptsize$[+0.024,+0.332]$} & 10/12 $>0$ \\
Restore: norm-matched random donor & $+0.124$ & {\scriptsize$[-0.065,+0.322]$} & 6/12 $>0$ \\
\midrule
\multicolumn{4}{l}{\emph{Matched all-heads controls at 2{,}000 tokens ($\Delta$ vs.\ own base)}} \\
\quad Disclosure span, target document, query window & $-0.092^{*}$ & {\scriptsize$[-0.145,-0.046]$} & 12/12 \\
\quad Same slot, neutral document, query window & $+0.007$ & {\scriptsize$[-0.013,+0.029]$} & 6/12 \\
\quad Control span, target document, query window & $-0.039$ & {\scriptsize$[-0.982,+0.888]$} & 8/12 \\
\quad Disclosure span, target document, post-slot & $-0.105^{*}$ & {\scriptsize$[-0.157,-0.059]$} & 12/12 \\
\quad Same slot, neutral document, post-slot & $+0.003$ & {\scriptsize$[-0.024,+0.028]$} & 6/12 \\
\quad Control span, target document, post-slot & $-0.054$ & {\scriptsize$[-0.989,+0.870]$} & 7/12 \\
\quad net disclosure-specific (target $-$ neutral slot, query) & $-0.099^{*}$ & {\scriptsize$[-0.157,-0.046]$} & 11/12 \\
\quad channel comparison (removal $-$ directional attn.\ removal) & $+0.025$ & {\scriptsize$[-0.055,+0.100]$} & exact $p=0.54$ \\
\quad content-specific component (target $-$ neutral donor) & $+0.022$ & {\scriptsize$[-0.012,+0.056]$} & 8/12 $>0$ \\
\bottomrule
\end{tabular*}
\begin{tablenotes}
\footnotesize
\item \textit{Notes:} Attention weights from the decision computation to the listed span are zeroed and renormalized for the listed heads over the listed positions; all runs share one attention implementation so conditions differ from their base only by the intervention. The relevant scale is the disclosure's own contrast at 2k (target minus neutral, +0.22 decision-score units in these runs): the all-heads conditions shift the readout by roughly a quarter to a third of that contrast (11/12 firms in a consistent direction). The matched controls below complete the identification: the same blackout on the neutral document's slot is null, so the net disclosure-specific component---the T-document effect minus the neutral-slot sham---is 44--48\% of the disclosure contrast. The content-free control-span rows are unstable across firms (note their intervals) and serve as a failed control, not as part of the identification. Donor patches replace the selected heads' outputs over the decision-query window with the same positions' outputs from a 2k run; the content-specific component (last row) is small relative to the generic patch effect. The channel-comparison row differences, within firm, the recurrent-transplant removal against the directional attention removal (the negative of the matched signed difference-in-differences; wrong-direction firms enter with their sign): the two channel effects are statistically indistinguishable in size. Brackets: seeded firm bootstrap, 95\%.
\end{tablenotes}
\end{threeparttable}
\end{table}

\begin{table}[p]
\centering
\begin{threeparttable}
\caption{Matched decomposition of extract-then-decide: all cells}
\label{tab:app_matched_full}
\footnotesize
\setlength{\tabcolsep}{3pt}
\begin{tabular*}{\textwidth}{@{\extracolsep{\fill}}lcccc}
\toprule
Condition & 2k, before & 2k, after & 128k, before & 128k, after \\
\midrule
Model's own targeted extraction & \begin{tabular}{@{}c@{}}$+0.126^{*}$\\{\scriptsize$[+0.077,+0.174]$}\end{tabular} & \begin{tabular}{@{}c@{}}$+0.113^{*}$\\{\scriptsize$[+0.075,+0.153]$}\end{tabular} & \begin{tabular}{@{}c@{}}$+0.085^{*}$\\{\scriptsize$[+0.067,+0.103]$}\end{tabular} & \begin{tabular}{@{}c@{}}$+0.055^{*}$\\{\scriptsize$[+0.038,+0.072]$}\end{tabular} \\
Experimenter-written targeted extraction & \begin{tabular}{@{}c@{}}$+0.165^{*}$\\{\scriptsize$[+0.104,+0.224]$}\end{tabular} & \begin{tabular}{@{}c@{}}$+0.151^{*}$\\{\scriptsize$[+0.106,+0.199]$}\end{tabular} & \begin{tabular}{@{}c@{}}$+0.088^{*}$\\{\scriptsize$[+0.074,+0.102]$}\end{tabular} & \begin{tabular}{@{}c@{}}$+0.056^{*}$\\{\scriptsize$[+0.042,+0.069]$}\end{tabular} \\
Another model's targeted extraction & \begin{tabular}{@{}c@{}}$+0.060^{*}$\\{\scriptsize$[+0.033,+0.086]$}\end{tabular} & \begin{tabular}{@{}c@{}}$+0.050^{*}$\\{\scriptsize$[+0.016,+0.082]$}\end{tabular} & \begin{tabular}{@{}c@{}}$+0.044^{*}$\\{\scriptsize$[+0.017,+0.072]$}\end{tabular} & \begin{tabular}{@{}c@{}}$+0.013$\\{\scriptsize$[-0.004,+0.030]$}\end{tabular} \\
Model's own verbatim copy of the passage & \begin{tabular}{@{}c@{}}$+0.020$\\{\scriptsize$[-0.013,+0.053]$}\end{tabular} & \begin{tabular}{@{}c@{}}$+0.016$\\{\scriptsize$[-0.013,+0.043]$}\end{tabular} & \begin{tabular}{@{}c@{}}$+0.047^{*}$\\{\scriptsize$[+0.029,+0.067]$}\end{tabular} & \begin{tabular}{@{}c@{}}$+0.018^{*}$\\{\scriptsize$[+0.009,+0.027]$}\end{tabular} \\
Verbatim re-presentation of the passage & \begin{tabular}{@{}c@{}}$+0.002$\\{\scriptsize$[-0.016,+0.023]$}\end{tabular} & \begin{tabular}{@{}c@{}}$-0.002$\\{\scriptsize$[-0.024,+0.024]$}\end{tabular} & \begin{tabular}{@{}c@{}}$+0.013^{*}$\\{\scriptsize$[+0.001,+0.029]$}\end{tabular} & \begin{tabular}{@{}c@{}}$+0.004$\\{\scriptsize$[-0.005,+0.011]$}\end{tabular} \\
\bottomrule
\end{tabular*}
\begin{tablenotes}
\footnotesize
\item \textit{Notes:} Mean Use with seeded firm-bootstrap 95\% intervals; $^{*}$ excludes zero. Conditions as in Table~\ref{tab:decomposition}.
\end{tablenotes}
\end{threeparttable}
\end{table}

\end{document}